\documentclass{article} 
\usepackage{iclr2023_conference,times}

\usepackage{amsmath,amsfonts,bm}

\def\eqref#1{equation~\ref{#1}}

\def\1{\bm{1}}

\DeclareMathAlphabet{\mathsfit}{\encodingdefault}{\sfdefault}{m}{sl}
\SetMathAlphabet{\mathsfit}{bold}{\encodingdefault}{\sfdefault}{bx}{n}

\usepackage{hyperref}
\usepackage{url}
\usepackage{hyperref}       
\usepackage{url}            
\usepackage{booktabs}       
\usepackage{amsfonts}       
\usepackage{nicefrac}       
\usepackage{microtype}      
\usepackage{xcolor}         
\usepackage{url}            
\usepackage{booktabs}       
\usepackage{amsfonts}       
\usepackage{nicefrac}       
\usepackage{microtype}      
\usepackage{subcaption}
\usepackage{amsmath,amsfonts,amssymb}
\usepackage{breqn}
\usepackage{tabularx}
\usepackage{multirow}
\usepackage{graphicx}
\usepackage{color}
\usepackage{tcolorbox}
\usepackage{CJK}
\usepackage{adjustbox}
\usepackage{xcolor}
\usepackage{colortbl}
\usepackage{multicol}
\usepackage{vwcol} 
\newsavebox{\algleft}
\newsavebox{\algright}
\usepackage{bm}
\usepackage{booktabs}
\usepackage{ctable} 
\usepackage{amsmath,stackengine}
\usepackage[linesnumbered,ruled,vlined]{algorithm2e}
\usepackage{times}
\usepackage{latexsym}
\usepackage{bbm}
\usepackage[utf8]{inputenc} 
\usepackage[T1]{fontenc}    
\usepackage{hyperref}       
\usepackage{url}            
\usepackage{booktabs}       
\usepackage{amsfonts}       
\usepackage{nicefrac}       
\usepackage{microtype}      
\usepackage{xcolor}         
\usepackage{dsfont}
\newcommand{\zixuan}[1]{{\color{red}{\small\bf\sf [zixuan: #1]}}}

\title{Continual Pre-Training of Language Models}

\author{
Zixuan Ke$^{1*}$, 
~Yijia Shao$^{2*}$,
~Haowei Lin$^{2}$\thanks{Equal contribution}~, 
~Tatsuya Konishi$^3$\thanks{The work was done when this author was visiting Bing Liu's group at University of Illinois at Chicago.}~, 
~Gyuhak Kim$^1$, and 
Bing Liu$^{1}$\thanks{Correspondance author. Bing Liu <\texttt{liub@uic.edu}>}\\ 
$^1$Department of Computer Science, University of Illinois at Chicago\\
$^2$Wangxuan Institute of Computer Technology, Peking University\\
$^3$KDDI Research\\
$^1$\texttt{\{zke4,gkim87,liub\}@uic.edu}\\  $^2$\texttt{\{shaoyj,linhaowei\}@pku.edu.cn} \\
$^3$\texttt{tt-konishi@kddi-research.jp}
}
\iclrfinalcopy 
\begin{document}

\maketitle



\begin{abstract}
{\color{black}{Language models (LMs) have been instrumental for the rapid advance of natural language processing. This paper studies continual {\color{black}pre-training} of LMs, in particular, \textit{continual domain-adaptive pre-training} (or \textit{continual DAP-training}). Existing research has shown that further pre-training an LM using a domain corpus to adapt the LM to the domain can improve the end-task performance in the domain.~This paper proposes a novel method to \textit{continually DAP-train} an LM with a sequence of unlabeled domain corpora to adapt the LM to these domains to improve their end-task performances.
The key novelty of our method is a soft-masking mechanism that directly controls the update to the LM. 
A novel proxy is also proposed to preserve the general knowledge in the original LM. Additionally, it contrasts the representations of the previously learned domain knowledge (including the general knowledge in the pre-trained LM) and the knowledge from the current full network 
to achieve knowledge integration. The method not only overcomes \textit{catastrophic forgetting}, but also achieves \textit{knowledge transfer} to improve end-task performances. 
Empirical evaluation demonstrates the effectiveness of the proposed method.}\footnote{The code is available at \url{https://github.com/UIC-Liu-Lab/ContinualLM}}}

\end{abstract}

\section{Introduction}
\label{sec:intro}









{\color{black}
Pre-trained language models (LMs) like BERT~\citep{DBLP:conf/naacl/DevlinCLT19} and RoBERTa~\citep{DBLP:journals/corr/abs-1907-11692} have significantly advanced NLP. Recently, LMs have also been used by many \textit{continual learning} (CL) systems to learn a sequence of end-tasks incrementally~\citep{ke2021achieving,sun2020lamol,huang2021continual}, which we call \textit{continual end-task learning}. {\color{black}It is also desirable to continually pre-train LMs themselves}. 
This includes (1) \textit{continual general pre-training}, which incrementally updates the LM using the most recent data that has a similar distribution as the pre-training data, and (2) \textit{continual domain-adaptive pre-training}, {\color{black}which 
further pre-trains a LM incrementally} 
to adapt it to a sequence of domains. Note that \textit{LM editing} (with or without continual learning) \citep{DBLP:conf/iclr/MitchellLBFM22} that corrects mistakes learned in the LM is {\color{black}a special case of continual end-task learning{\color{black}~\citep{kim2022theoretical}}
as each editing task or a group of editing tasks learned together is basically a \textit{task} in continual learning, which aims to perform the editings correctly without interfering with or forgetting the other knowledge already learned in the current LM}. 
} 

This paper focuses on \textit{continual \textbf{d}omain-\textbf{a}daptive \textbf{p}re-training} (or \textit{continual DAP-training}) of LMs. It is known that \textit{DAP-training}\footnote{{\color{black}Depending on different contexts or authors, DAP-training is also called \textit{post-training} or \textit{pre-finetuning}.}} an LM (without continual learning) using a large unlabeled domain corpus before end-task fine-tuning achieves better results~\citep{DBLP:conf/acl/GururanganMSLBD20,DBLP:conf/naacl/XuLSY19,ke2022dga}. This paper goes a step further to continually learn to improve an LM's ability to handle new or emerging domains or topics \textit{without} forgetting the skills or knowledge learned in the past. This is important in the real world, where the data shifts constantly and new domains, events or topics keep emerging  \citep{ke2022dga} and the LM needs to be updated to serve the users better. 


{\color{black} We call this problem \textit{continual DAP-training}. 
Starting from a pre-trained general LM (i.e., the LM has already been pre-trained on $D_0$), we incrementally DAP-train a sequence of domain corpora $D_1, D_2, ...$. Once a domain is trained, its data is no longer accessible. This is different from conventional \textit{continual learning} (\textbf{CL}) where each task is an end-task. In the proposed continual DAP-training, each task is an unlabeled domain corpus to be learned. An end-task fine-tunes {\color{black}the continually DAP-trained LM to evaluate its performance}. It is worth noting that $D_0$ is usually a \textit{broad or general domain} (e.g., News). In practice, a continually DAP-trained LM may be trained by individual users, institutions or a mix of both who have one or more large corpora of some particular domains. In such cases, the raw data may not be shared, but the final LM can be shared by all. 

}



There are multiple desiderata for a continual DAP-training system: 
\textbf{(1)} It should not suffer from \textit{catastrophic forgetting} \textbf{(CF}),
 i.e., it should perform reasonably well on learned domains. This requires the system (a) to overcome CF for each new domain and (b) to overcome CF for the general language knowledge in the LM. This is important because 
the knowledge learned from each domain alone will not be sufficient for good end-task performances. 
\textbf{(2)} It should encourage knowledge transfer (\textbf{KT}) across domains to achieve improved end-task performances.
This requires the system to enable (a) \textit{forward transfer}, learning a new domain by leveraging the knowledge from previous domains, and (b) \textit{backwards transfer}, gaining improved performance on previous domains after learning a relevant new domain.
\textbf{(3)} It should work without requiring the domain-ID for each end-task fine-tuning.


None of the existing CL methods can achieve all the above. This paper represents a step towards achieving them. The proposed method is called \textbf{DAS} (Continual \textit{\textbf{DA}}-pre-training of LMs with \textit{\textbf{S}}oft-masking).
DAS proposes a novel \textit{soft-masking mechanism} that computes the importance (a real number between 0 and 1) of units\footnote{For simplicity, we use the term \textit{units} to mean both \textit{attention heads} and \textit{neurons}.} for general or domain knowledge and soft-mask them based on their importance values to control the backward gradient flow. In the forward pass, soft-masking is not applied, which encourages KT across domains. It does not isolate any sub-network for any domain so that the knowledge in the full LM can be leveraged for end-task fine-tuning. 

To apply this mechanism, DAS implements two functions: (1) \textit{Initialization}, which computes the importance of units to \textbf{the general knowledge} in the LM without accessing the LM pre-training data ($D_0$). It is applied on the pre-trained LM before the continual learning starts, and (2) \textit{continual learning}, which DAP-trains each domain while preventing CF on the general and domain knowledge and encouraging cross-domain KT.
In (1), it is not obvious how to compute the importance without pre-training data. DAS proposes a novel proxy based on \textit{robustness} to compute the importance of units for the general knowledge. In (2), the soft-masking is directly applicable because we have the domain data and the importance can be computed based on its gradient inspired by the pruning community~\citep{li2021differentiable,michel2019sixteen}. Moreover, DAS contrasts the previously learned knowledge and the full (including both the learned domains and the current domain) knowledge to encourage the current domain representation to learn knowledge that is not already in the knowledge learned from previous domains and integrate it with the learned knowledge\footnote{Contrasting the past domains and only the domain-specific knowledge gives poorer results (see Sec.~\ref{sec:results}) as it causes the two types of knowledge to split rather than to integrate.}. In end-task fine-tuning, DAS does not requires the domain-ID as all knowledge is accumulated into the DAP-trained LM.



In summary, this work makes the following contributions. 
\textbf{(i)} It studies the new problem of \textit{continual DAP-training} and  
discovers that the full LM is needed for a good continual DAP-training method. The popular \textit{parameter-isolation approach} to overcoming CF in convention CL is unsuitable.
\textbf{(ii)} It proposes a novel soft-masking method to overcome CF and to encourage KT, and a constrative learning based method for knowledge integration.  
\textbf{(iii)} To preserve the general knowledge in the LM, a novel proxy is also proposed.
\textbf{(iv)} Experimental results demonstrate the effectiveness of DAS. 

{\color{black}
\section{Related Work}
\label{sec:related}
\textbf{DAP-training.} DAP-training 
can be achieved by directly updating the LM \citep{DBLP:conf/naacl/XuLSY19,sun2019fine,lee2020biobert,alsentzer2019publicly,DBLP:conf/acl/GururanganMSLBD20,chakrabarty2019imho,ke2022dga} or by training only a small set of additional parameters. For example, \cite{pfeiffer2020adapterfusion,wang2020k,ke2021achieving,ke2021Classic,ke2021adapting} trained adapters and \cite{gu2021ppt} trained a prompt to adapt to a domain. While adapter and prompt could be effective, transfer knowledge among these additional modules is usually challenging and can be inaccurate. DAS belongs to the former family that directly updates the LM. This is very challenging for CL due to CF. To our knowledge, no existing system in this family is about CL. }

\noindent
\textbf{Continual learning.} 
Most CL methods were proposed to overcome CF: 
(1) \textit{Regularization methods}~\citep{Kirkpatrick2017overcoming,Seff2017continual} compute the importance of each parameter to previous tasks and uses a regularizer to penalize the sum of changes.
DAS is related to but also very different from 
EWC~\citep{Kirkpatrick2017overcoming}. 
(1) DAS does not control each parameter/weight, but only attention heads or neurons based on their importance scores. This gives less forgetting (see the forgetting rate in Table~\ref{tab:dapt_result}) because even a small change to each parameter for a neuron can give a large total change to the neuron's activation. (2) DAS directly controls the backward gradient flow on each neuron, which is more fine-grained and effective than the sum of changes of all parameters. Our experimental results confirm that EWC is significantly poorer than DAS (see Table~\ref{tab:dapt_result}). (2) \textit{Replay methods} retain~\citep{Rebuffi2017,wang2020efficient} or generate some data of old tasks~\citep{Shin2017continual,He2018overcoming} and use them in learning a new task;
(3) \textit{parameter-isolation methods} \citep{Serra2018overcoming,wortsman2020supermasks} allocate neurons and parameters or sub-networks for different tasks/domains and mask them in task learning. 
For continual DAP-training, 
this means that end-tasks cannot use the general knowledge in the LM, which results in poor end-task performances.

In NLP, CL has been used for slot filling~\citep{shen-etal-2019-progressive}, language learning~\citep{li2019compositional},
sentiment analysis~\citep{ke2021achieving}, topic modeling~\citep{gupta2020neural}, question answering~\citep{greco2019psycholinguistics} and text classification~\citep{sun2020lamol,huang2021continual,chuang2020lifelong}. But none is for DAP-training.~Some recent CL papers concern LMs. 
The system in \citep{madotto2020continual} learns separate adapters for different domains and thus has no CF or KT. 
DEMIX~\citep{gururangan2021demix} 
initializes the new adapter with the closest old adapter.  
{\color{black} 
CPT~\citep{ke2022cpt} and ELLE~\citep{DBLP:conf/acl/QinZLL0SZ22} are most closely related to DAS. However, CPT uses the parameter-isolation approach to learn and protect each task, which is weak (see Sec. \ref{sec:results}). It also needs domain-ID in end-task fine-tuning. ELLE has to start from pre-training the LM itself rather than from a pre-trained LM like DAS. It also uses a large memory (1G per domain) to store the replay data (including the pre-training data) and expands the network for each domain. Neither is required in DAS. 
\citet{jin2021lifelong} evaluated several existing CL techniques in a similar setting as DAS and performed analyses on dealing with CF. However, no new technique was proposed in the paper.}

\noindent
\textbf{Neural network pruning.}
Many parameters in a network are redundant and can be pruned~\citep{li2021differentiable,lai2021parp,
michel2019sixteen,voita2019analyzing}. 
Existing methods include discarding parameters with small absolute values~\citep{han2015learning,guo2016dynamic}, accumulated gradient \citep{michel2019sixteen}, 
and lottery ticket hypothesis \citep{brix-etal-2020-successfully}. However, these methods are not directly applicable as  we need to preserve not only individual domain knowledge but also the general knowledge in the LM. For general knowledge, since we do not have any pre-training data, a proxy is proposed based on robustness.  For domain knowledge, we adopt a pruning method but use the importance as soft-masks as we want to accumulate knowledge rather than to compress the LM.

\textbf{Contrastive Learning.} 
Contrastive learning~\citep{chen2020simple,he2020momentum} learns good 
representations by maximizing the similarity of positive pairs and minimizes that of negative pairs, 
{\color{black}
\begin{equation}
\label{eq.relate_contrast}
    \mathcal{L}_{\text{contrast}} = -\frac{1}{N}\sum_{n=1}^{N}\log\frac{e^{(\text{sim}(q_n, q^+_n)/\tau)}}{\sum_{j=1}^{N}e^{(\text{sim}(q_n,q^+_j)/\tau)}},
\end{equation}
}
where $N$ is the batch size, $\tau$ is a temperature parameter, $\text{sim}(\cdot)$ is a similarity metric, and $q_n$ and $q^+_n$ are representations for positive pairs $x_n$ and $x_n^+$. DAS contrasts the learned knowledge from previous domains and the pre-trained LM (general knowledge) with the full knowledge (including both the previous domains and current domain knowledge) to achieve a complementary effect. 

\section{Proposed DAS Technique}
\label{Sec: preliminary}











\begin{figure*}[t!]
\centering
\includegraphics[width=0.7\textwidth]{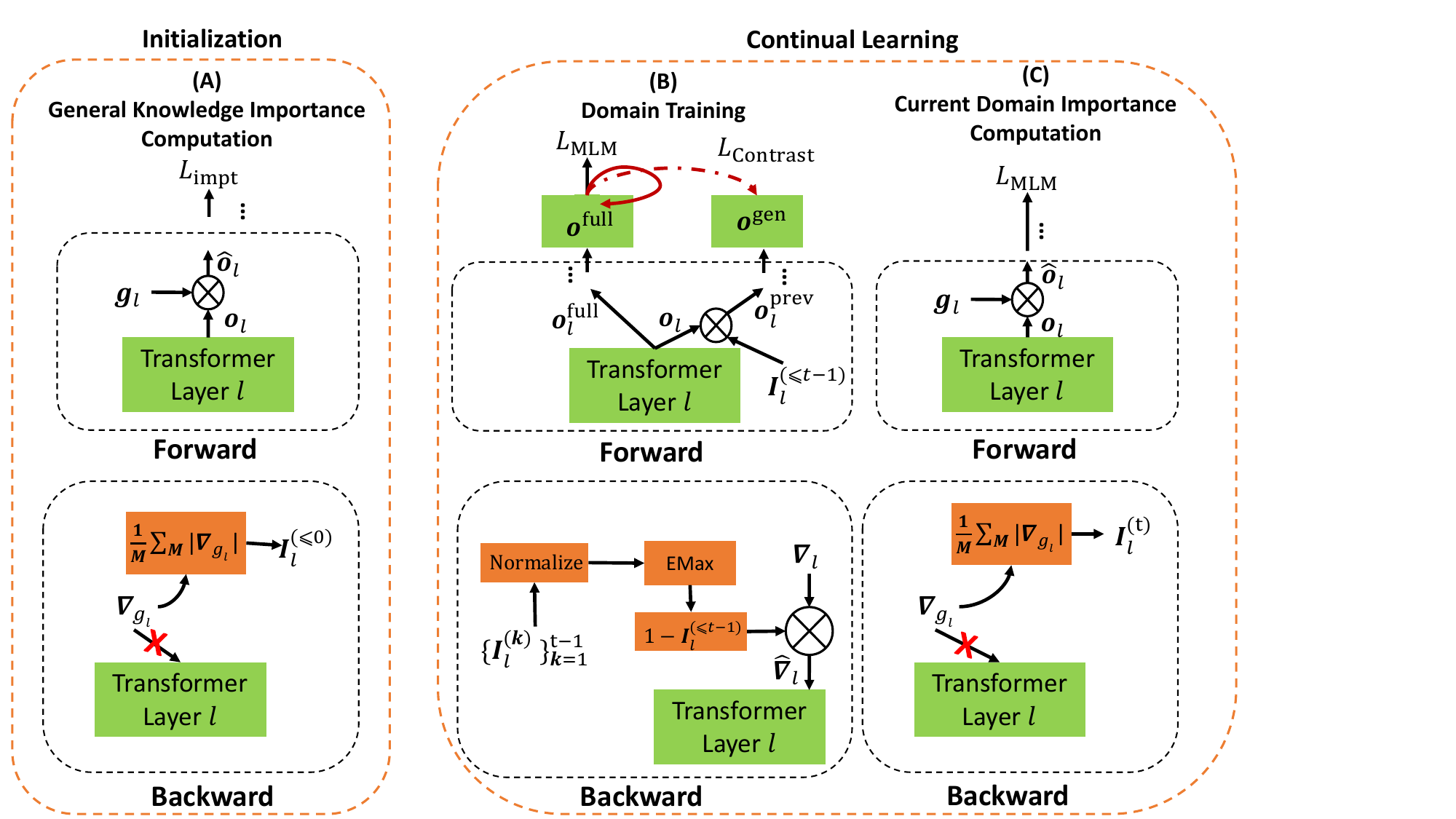}
 \vspace{-2mm}
\caption{
{\color{black}Illustration of DAS. {\color{black}The red cross indicates that the gradient is not used to update the Transformer but only to compute importance.} 
(A) \textbf{Initialization} (Sec.~\ref{sec:initialization}) computes the importance of units 
for the general knowledge in the LM. (B) \textbf{Domain Training} (Sec.~\ref{sec:training}) trains a new domain using the importance scores as soft-masks and contrasts the previously learned knowledge
and the full knowledge.
(C) \textbf{Importance Computation} (Sec.~\ref{sec:after_training}) computes the importance of the units for the current domain.
}
}
\label{fig:overview}
\vspace{-4mm}
\end{figure*}

Continual DAP-training in DAS is based on two main ideas: (1) preserving the important general language knowledge in the LM and the knowledge learned from previous domains to overcome CF by soft-masking units based on their importance, which also facilitates cross-task knowledge transfer (KT), and (2) encouraging the model to learn complementary representations of the current domain and previous domains to achieve knowledge integration.
Figure~\ref{fig:overview} gives an overview of DAS. 

{The whole learning consists of two main functions: (i) \textbf{\textit{initialization}} and (ii) \textbf{\textit{continual learning}}. (i) computes the importance of units to the general language knowledge in the LM. 
It is done before the continual learning starts. (ii) is for continual learning, which consists of two steps: (a) \textbf{\textit{domain training}} and (b) \textbf{\textit{importance computation}}. (a) takes the importance scores accumulated so far (including those to the general knowledge in the original LM and to the knowledge learned from previous domains) and the input data of the current domain to learn the domain and to achieve (1) and (2) above, while (b) computes the importance scores for the current domain for future use. The following sub-sections present each function and step in detail.
}

\subsection{Initialization: Computing Importance of Units to the General Knowledge}
\label{sec:initialization}

This initialization function computes the importance of units (attention heads and neurons) in the Transformer for the \textit{general} knowledge in the original LM. The key components of a Transformer are \textit{multi-head attention layer}, \textit{intermediate layer} and \textit{output layer}. {Below, we use ``layer'' or $l$ to indicate any of these three layers because our method treats the three layers similarly.}

\textbf{Importance of units in a layer.} It has been found that not all units in a layer are important \citep{michel2019sixteen}. We introduce a {\color{black}\textit{virtual parameter}}, $\bm{g}_{l}$, for computing the importance of the units in a layer $l$. {\color{black}We call these \textit{virtual} parameters as each $g^{(k)}$ is initialized to 1. We only need the gradient on each parameter to compute the importance of its corresponding unit, no update to any parameter. }
\begin{equation}
\label{eq:gmhatt}
\hat{\bm{o}}_l=\bm{g}_{l} \otimes \bm{o}_l,
\end{equation}
where $\bm{o}_l$ refers to the output of layer $l$ (which can be any of the three layers mentioned above). The $\otimes$ refers to element-wise multiplication, i.e., each variable $g_{l,i}$ in $\bm{g}_l$ corresponding to a unit (a neuron or attention head) in the layer.  We adapt the gradient-based importance detection method in \citep{michel2019sixteen} for our purpose. Given a dataset $D=\{(\bm{x}_n,{y}_n)\}_{n=1}^N$ of $N$ samples  ($y_n$ is the class label of $\bm{x}_n$ as \citep{michel2019sixteen} worked on supervised learning), the importance of neurons or heads in the layer is estimated with a gradient-based proxy score
\begin{equation}
\label{eq:importance}
\bm{I}_{l} = \frac{1}{N}\sum_{n=1}^N|\frac{\partial\mathcal{L}_{\text{impt}}(\bm{x}_n,{y}_n))}{\partial \bm{g}_{l}}|,
\end{equation}
where $\mathcal{L}_{\text{impt}}$ is a task-specific loss function. Note the virtual parameter $\bm{g}_{l}$ is initialized as all 1's, and is not changed. This is because we need only its average gradient $\nabla_{\bm{g}_l}$ (the term within $||$ in Eq.~\ref{eq:importance}) over all the data to compute the importance and will not use the gradient to update the virtual parameter. In training (Sec.~\ref{sec:training} and Fig~\ref{fig:overview} (B)), the virtual parameter can be discarded. {\color{black}The resulting $\bm{I}_{l}$ is of the same size as $\bm{g}_{l}$, each entry corresponding to the importance of a unit (a neurons or attention head).}

Recall that the \textit{initialization} function is to learn the importance of units to the \textit{general knowledge} in the LM ({\color{black}denoted as $\bm{I}_l^{(0)}$}). Although Eq.~\ref{eq:importance} offers a possible way, 
it is not directly applicable. If we use the domain data at hand and employ the MLM loss as $\mathcal{L}_{\text{impt}}$, $\nabla_{\bm{g}_{l}}$ only gives the importance for the \textit{domain-specific} knowledge. However, to compute the importance of units to the general knowledge in the LM (which is our goal), we need the original data used in pre-training the LM to compute the $\mathcal{L}_{\text{impt}}$. In practice, such data is not accessible to users of the LM. Further, label is needed in Eq.~\ref{eq:importance} but our domain corpus is unlabeled in DAP-training.
To address these issues, we propose a \textit{proxy KL-divergence loss} ($\mathcal{L}_{\text{proxy}}$) to replace $\mathcal{L}_{\text{impt}}$ to learn units importance for the general knowledge.

\textbf{Proxy KL-divergence loss.} 
We propose to use model \textit{robustness} as the proxy, i.e., we try to detect units that are important for LM's \textit{robustness}. Their gradients, $\nabla_{\bm{g}_{l}}$, then indicate the robustness and the importance to the LM model. Our rationale is as follows: If an $I_{l,i}^{(0)}$ (the importance of unit $i$ in layer $l$) has a high value, then it is important to the LM's robustness because its change can cause the LM to change a lot. It is thus an important unit. In contrast, if $I_{l,i}^{(0)}$ is small, it is a less important unit.
To compute the \textit{robustness} of the LM, we take a subset of the current domain data $\{\bm{x}^{\text{sub}}_n\}$\footnote{{\color{black}We use a subset to save computation as we assume that the DAP-training domain can be very large. In Sec.~\ref{sec:experiments}, we show that a subset is sufficient to compute the importance of units for the general knowledge.}} 
(no label in DAP-training)
and input $\bm{x}^{\text{sub}}_n$ twice to the LM  {\color{black}to obtain two representations of it and then compute the KL-divergence between them}, 
\begin{equation}
\mathcal{L}_{\text{impt}} = \text{KL}(f^1_{\text{LM}}(\bm{x}^{\text{sub}}_n),f^2_{\text{LM}}(\bm{x}^{\text{sub}}_n)),
\label{eq:proxy}
\end{equation}
where $f_{\text{LM}}^1$ and $f_{\text{LM}}^2$ are the LM with different dropout masks. We don't need to add any additional dropouts to implement these two as 
the Transformer already has dropout masks placed on fully-connected layers and attention probabilities. Thus, simply feeding the same input to the Transformer twice will get two representations with different dropout masks. Since dropout is similar to adding noise, the difference between the two representations can be regarded as the \textit{robustness} of the LM. 

\subsection{Training: Learning a New Domain via Soft-masking and Contrastive Loss}
\label{sec:training}


Recall we want to preserve the learned knowledge in the LM during DAP-training using the accumulated importance $\bm{I}_{l}^{(\le t-1)}$ when we learn domain $t$, which includes both the importance for the general knowledge $\bm{I}_l^{(0)}$ (Sec.~\ref{sec:initialization}) and learned domain-specific knowledge $\bm{I}_l^{(k)}$ of each domain $k$ ($k$ can be any domain in $\{1...t-1\}$) that has been learned (Sec.~\ref{sec:after_training}). This is achieved by soft-masking the learning based on accumulated importance as follows.\footnote{{\color{black}Before training, we normalized the the importance values in each layer $l$ for a domain $k$ by making the importance scores for all units in the layer having a mean of 0 and standard deviation of 1. To further facilitate soft-masking, the normalized importance scores are rounded by a \texttt{Tanh} activation so that the values are in the interval of [0,1]. To simplify the notation, we still use $\bm{I}^{(k)}_{l}$ to represent the resulting importance.}}

\textbf{Accumulating importance.} We accumulate the importance after task $t-1$ was learned is done {\color{black}via element-wise max (EMax)} as follows: 
\begin{equation}
\bm{I}_{l}^{(\le t-1)} = \text{EMax}(\{\bm{I}_{l}^{(t-1)},\bm{I}^{(\le t-2)}_{l}\}),
\label{eq.accumulate}
\end{equation}
where $t$ refers to the current task-ID and $\bm{I}^{(\le t-2)}_{l}$ refers to the previously accumulated importance at task $t-2$. 
We do not need to save $\bm{I}_{l}^{0}$ and all $\{\bm{I}^{(k)}_{l}\}_{k=1}^{t\text{-1}}$ for Eq. \ref{eq.accumulate}. We only save the incrementally accumulated importance after training of each task. 

\textbf{Soft-masking units.} Given the accumulated importance $\bm{I}_{l}^{(\le t-1)}$ of layer $l$ and the DAP-training loss $\mathcal{L}_{\text{DAP-train}}$ (typically the MLM loss; we also propose an additional loss in Eq.~\ref{eq:contrast}), we constrain (or soft-mask) its corresponding gradient ($\nabla_l$) flow as follows, 
\begin{equation}
\label{eq.softmask}
\hat{\nabla}_{l} = (1-\bm{I}_{l}^{(\le t-1)}) \otimes \nabla_{l},
\end{equation}
As mentioned in Sec.~\ref{sec:initialization}, we expand (by copying) the importance $\bm{I}_{l}^{(\le t-1)}$ to match the dimensions of $\nabla_l$ to apply it to all associated parameters. This is \textit{soft-masking} as each element in $\bm{I}_{l}^{(\le t-1)}$ is a real number in $[0,1]$ (not binary \{0, 1\}), which gives the model the flexibility to adjust any unit. 


We note that the above soft-masks are only applied in the backward pass, but not in the forward pass, which encourages knowledge transfer as each domain training can leverage the knowledge learned from all past domains. To further encourage the model to learn a good representation from both the accumulated knowledge ($\bm{I}^{(\le t-1)}_{l}$) and the full knowledge (both accumulated and current domain knowledge), we introduce a contrastive learning method to encourage complementary representation.


\noindent\textbf{Integrating the previously learned knowledge and the current domain knowledge.}
{\color{black}Soft-masking helps prevent forgetting the previously learned knowledge. We want to further encourage knowledge transfer by integrating the new and learned knowledge. We propose to contrast the previously learned knowledge and the full knowledge (both previously learned knowledge and the current domain knowledge). Note that the contrasting cannot do anything to the shared past knowledge as it is protected by soft-masks. Thus, it effectively pushes the current domain knowledge away to be complementary to the past knowledge. This is done based on the current domain data as follows.}

{\color{black}\textbf{\textit{Contrasting the learned and full knowledge}.} We denote the output of LM without any consideration of importance as $\bm{o}^{\text{full}}$, which refers to the full knowledge. We further denote the output of LM that is multiplied by the importance (i.e., $\bm{I}^{(\le t-1)}_{l} \otimes \bm{o}_l$) as $\bm{o}^{\text{prev}}$, which refers to the previously learned knowledge. We contrast the two by using $\bm{o}^{\text{full}}$ as anchor and $\bm{o}^{\text{full}}$ with different dropouts as positive samples (dentoed as $\bm{o}^{\text{full+}}$).  $\bm{o}^{\text{prev}}$ is used as negative instances.

Formally, with $\bm{o}_n^{\text{full}}$, $\bm{o}_n^{\text{full+}}$, and $\bm{o}_n^{\text{prev}}$, our contrastive loss is ($\text{sim}(\cdot)$ is the cosine similarity), {\color{black}
\vspace{-1mm}
\begin{equation}
\label{eq:contrast}
\vspace{-1mm}
\small
\mathcal{L}_{\text{contrast}} =  -\frac{1}{N}\sum_{n=1}^{N}\text{log}\frac{e^{\text{sim}(\bm{o}_n^{\text{full}},\bm{o}_n^{\text{full+}})}/\tau}{\sum_{j=1}^{N}(e^{\text{sim}(\bm{o}_n^{\text{full}},\bm{o}_j^{\text{full+}})/\tau}+e^{\text{sim}(\bm{o}^{\text{full}}_n,\bm{o}_j^{\text{prev}})/\tau})}.
\end{equation}}
Compared to Eq.~\ref{eq.relate_contrast},
the second term is added in the denominator, i.e., representations in the previously learned knowledge as additional negative instances. Figure~\ref{fig:overview} (B) shows a red arrow pointed from $\bm{o}^{\text{full}}$ to itself, indicating the positive instances are from inputting twice. The dashed red arrow pointing to $\bm{o}^{\text{prev}}$ indicates the negative instances contrasting the full and previously learned knowledge.
}


\noindent
\textbf{Final Loss Function}.
The final DAP-training loss combines the Masked Language Model (MLM) loss after applying the proposed soft-masking for the general knowledge 
(Sec.~\ref{sec:initialization}) and the proposed contrastive loss ({\color{black}$\lambda$ is a hyper-parameter}),
\begin{equation}
\label{eq:final_loss}
\mathcal{L}_{\text{DAP-train}} = \mathcal{L}_{\text{MLM}} + \lambda\mathcal{L}_{\text{contrast}}
\end{equation}
\subsection{Compute importance of units to The current domain}
\label{sec:after_training}
\vspace{-1mm}
After training the new/current domain $t$, we learn the units importance by applying Eq.~\ref{eq:importance} for the domain. We do not need any proxy to compute  $\mathcal{L}_{\text{impt}}$ as in Eq. \ref{eq:proxy} because we can directly use the current domain data. Specifically, we randomly sample a subset (a hyper-parameter) of the current domain data $\{(\bm{x}^{\text{sub}}_n,y^{\text{sub}}_n)\}$, where $\bm{x}^{\text{sub}}_n$ is the input and $y^{\text{sub}}_n$ is the masked token as in MLM self-supervised loss. We can then easily compute the importance $\bm{I}^{(t)}_{l}$ by plugging $\mathcal{L}_{\text{MLM}}$ into $\mathcal{L}_{\text{impt}}$ in Eq. \ref{eq:importance}. The resulting $\bm{I}^{(t)}_{l}$ will be used in the next task by accumulating with the previously accumulated importance (Eq. \ref{eq.accumulate}) and soft-masking the learning (Eq. \ref{eq.softmask}). 



\begin{table*}[]
\centering
\resizebox{\textwidth}{!}{
\begin{tabular}{ccc|ccccc}
\specialrule{.2em}{.1em}{.1em}
\multicolumn{3}{c|}{Unlabelde Domain Datasets} & \multicolumn{5}{c}{End-Task Classification Datasets} \\
Source & Dataset & Size & Dataset & Task & \#Training & \#Testing & \#Classes \\
\specialrule{.1em}{.05em}{.05em}
\multirow{3}{*}{Reviews} & Yelp Restaurant & 758MB & Restaurant & Aspect Sentiment Classification (ASC) & 3,452 & 1,120 & 3 \\
 & Amazon Phone & 724MB & Phone & Aspect Sentiment Classification (ASC) &  239 & 553 & 2 \\
 & Amazon Camera & 319MB & Camera & Aspect Sentiment Classification (ASC) & 230 & 626 & 2 \\
 \hline
\multirow{3}{*}{Academic Papers} & ACL Papers & 867MB & ACL & Citation Intent Classification & 1,520 & 421 & 6 \\
 & AI Papers & 507MB & AI & Relation   Classification & 2,260 & 2,388 & 7 \\
 & PubMed Papers & 989MB & PubMed & Chemical-protein Interaction Prediction & 2,667 & 7,398 & 13 \\
\specialrule{.1em}{.05em}{.05em}
\end{tabular}
}
\vspace{-3mm}
\caption{
Statistics of datasets for DAP-training. {More details of }their end-task supervised learning datasets are given in Appendix~\ref{ap:dataset}. 
} 
\label{tab:dataset}
\vspace{-4mm}
\end{table*}

\section{Experiments}
\label{sec:experiments}


We use RoBERTa \citep{DBLP:journals/corr/abs-1907-11692}\footnote{\url{https://huggingface.co/roberta-base}} as the LM. Following the standard evaluation setup~\citep{DBLP:journals/corr/abs-1909-08383} and, after a domain is trained, its training data is discarded. After all domains are incrementally learned, the final model is evaluated by fine-tuning the end-tasks in all domains. 




\vspace{-1mm}
\subsection{Datasets and Baselines}
\label{sec:data-baselines}
\vspace{-1mm}
\textbf{Datasets:} 
Table \ref{tab:dataset} shows the statistics of the 6 \textit{unlabeled domain corpora} for DAP-training and their 6 corresponding \textit{end-task classification datasets}.\footnote{We down-sampled the \textit{PubMed} due to its huge original size. In general, our datasets are much smaller than those used in  \citep{DBLP:conf/acl/GururanganMSLBD20} (which used more than 11GB of data for each domain). Our experiments showed that a smaller dataset is sufficient and more data does not help. It also requires less computing power.} 3 of them are about reviews: \textit{Yelp Restaurant} \citep{DBLP:conf/naacl/XuLSY19}, 
\textit{Amazon Phone} \citep{DBLP:conf/emnlp/NiLM19}, \textit{Amazon Camera} \citep{DBLP:conf/emnlp/NiLM19}; 3 of them are academic papers: \textit{ACL Papers} \citep{DBLP:conf/acl/LoWNKW20}, \textit{AI Papers} \citep{DBLP:conf/acl/LoWNKW20}, and \textit{PubMed Papers}\footnote{\url{https://pubmed.ncbi.nlm.nih.gov/}}.
Their corresponding \textit{end-task classification datasets} are:\footnote{Our results are different from those presented in Table 5 of \citep{DBLP:conf/acl/GururanganMSLBD20} because we observe very high variances due to very small test sets and thus enlarge the test set and reduce the training set slightly.} \textit{Restaurant}\footnote{\url{https://alt.qcri.org/semeval2014/task4/}}, 
\textit{Phone} \citep{ding2008holistic,hu2004mining}, \textit{Camera} \citep{ding2008holistic,hu2004mining}, 
\textit{ACL} (ACL-ARC in \citep{DBLP:journals/tacl/JurgensKHMJ18}), \textit{AI} (SCIERC in \citep{DBLP:conf/emnlp/LuanHOH18}), and PubMed (CHEMPORT in \citep{kringelum2016chemprot}).  

\noindent
\textbf{Baselines.} We use 16 baselines, including non-continual learning (Non-CL) and continual learning (CL) baselines. {\color{black}All CL baselines are originally for learning supervised data except DEMIX.} We adapt them and replace their backbone with RoBERTa. Details of each baseline is given in Appendix~\ref{ap:baseline}. 

\textbf{Non-CL Baselines}: Each baseline here builds a separate model for each
task.  {\color{black}\textbf{(1) Pool. } We pool the data of all domains together and train only one model for all domains. }
\textbf{(2) RoBERTa}~\citep{DBLP:journals/corr/abs-1907-11692} uses RoBERTa for end-task fine-tuning without DAP-training. \textbf{(3) DAP-RoBERTa} uses the existing DAP-training method (MLM) in~\citep{DBLP:conf/acl/GururanganMSLBD20} to posst-train each domain separately. \textbf{(4) DAP-Adapter} adds adapter
layers in Transformer for each domain for DAP-training~\citep{jang2021continual,madotto2020continual,Houlsby2019Parameter}. Only the added adapters are trainable. In end-task fine-tuning, both RoBERTa and the adapters are trainable.  \textbf{(5) DAP-Prompt} is from 
\citep{DBLP:conf/emnlp/LesterAC21}. In DAP-training, RoBERTa (the LM) is fixed and only the prompts are trained. In end-task fine-tuning, both the LM and the trained prompt are trainable. 

\textbf{CL Baselines}: We use 2 naive baselines, which keep learning more domains with no mechanism to deal with CF or transfer. \textbf{(6) NCL} (Naive CL) continually DAP-trains the RoBERTa; and \textbf{(7) NCL-Adapter} continually DAP-trains a set of adapters~\citep{Houlsby2019Parameter}. 

8 baselines are CL systems: \textbf{(8) DEMIX}~\citep{gururangan2021demix} adds a new adapter for each new domain and initializes it with a previous adapter nearest to the new domain; 
\textbf{(9) BCL}~\citep{ke2021adapting} uses capsule networks. 
\textbf{(10) CLASSIC}~\citep{ke2021Classic} uses contrastive learning. \textbf{(11) KD} is knowledge distillation~\citep{hinton2015distilling}. 
\textbf{(12) EWC}~\citep{buzzega2020dark} is a popular
regularization-based method. \textbf{(13) DER++}~\citep{buzzega2020dark} is a replay method based on knowledge distillation. 16.4K tokens are saved for each domain in the replay memory, which is the largest memory we can use for the system to run. 
\textbf{(14) HAT}~\citep{Serra2018overcoming} is an effective \textit{parameter-isolation} method. HAT is applied to Transformer layers (i.e., self-attention, intermediate and output layers). {\color{black}\textbf{(15) HAT-All} is a HAT variant that uses all features \textit{from the LM} to do end-tasks (instead of only features from its domain sub-network as in HAT).} \textbf{(16) HAT-Adapter}~\citep{ke2021adapting} uses HAT within adapters.  ELLE~\citep{DBLP:conf/acl/QinZLL0SZ22} is not included as we adapted it for our purpose by learning from RoBERTa, {but it fails to converge.}  


\begin{table*}[]
\centering
\resizebox{\textwidth}{!}{
\begin{tabular}{cc|ccccccccccccccc}
\specialrule{.2em}{.1em}{.1em}
\multirow{2}{*}{Category} & Domain & \multicolumn{2}{c}{Restaurant} & \multicolumn{2}{c}{ACL} & \multicolumn{2}{c}{AI} & \multicolumn{2}{c}{Phone} & PubMed & \multicolumn{2}{c}{Camera} & \multicolumn{2}{c}{Average} & \multicolumn{2}{c}{Forget R.} \\
 & Model & MF1 & Acc & MF1 & Acc & MF1 & Acc & MF1 & Acc & MF1 & MF1 & Acc & MF1 & Acc & MF1 & Acc \\
\specialrule{.1em}{.05em}{.05em}
\multirow{5}{*}{Non-CL} &{\color{black}Pool} &{\color{black}80.96 } &{\color{black}87.80 } &{\color{black}69.69 } &{\color{black}74.11 } &{\color{black}68.55 } &{\color{black}75.97 } &{\color{black}84.96 } &{\color{black}86.95 } &{\color{black}73.34 } &{\color{black}86.03 } &{\color{black}90.83 } &{\color{black}77.25 } &{\color{black}81.50 } &\multicolumn{2}{c}{---}\\
\cline{2-17}
& RoBERTa & 79.81 & 87.00 & 66.11 & 71.26 & 60.98 & 71.85 & 83.75 & 86.08 & 72.38 & 78.82 & 87.03 & 73.64 & 79.27 & \multicolumn{2}{c}{---} \\
 & DAP-RoBERTa & 80.84 & \textbf{87.68} & 68.75 & 73.44 & 68.97 & 75.95 & 82.59 & 85.50 & 72.84 & 84.39 & 89.90 & 76.40 & 80.89 & \multicolumn{2}{c}{---} \\
 & DAP-Adapter & 80.19 & 87.14 & 68.87 & 72.92 & 60.55 & 71.38 & 82.71 & 85.35 & 71.68 & 83.62 & 89.23 & 74.60 & 79.62 & \multicolumn{2}{c}{---} \\
 & DAP-Prompt & 79.00 & 86.45 & 66.66 & 71.35 & 61.47 & 72.36 & 84.17 & 86.53 & \textbf{73.09} & 85.52 & 90.38 & 74.98 & 80.03 & \multicolumn{2}{c}{---} \\
\specialrule{.1em}{.05em}{.05em}
\multirow{11}{*}{\multirow{2}{*}{\begin{tabular}[c]{@{}c@{}}CL\\ DAP-train\end{tabular}}} & NCL & 79.52 & 86.54 & 68.39 & 72.87 & 67.94 & 75.71 & 84.10 & 86.33 & 72.49 & 85.71 & 90.70 & 76.36 & 80.77 & 1.14 & 1.05 \\
 & NCL-Adapter & 80.13 & 87.05 & 67.39 & 72.30 & 57.71 & 69.87 & 83.32 & 85.86 & 72.07 & 83.70 & 89.71 & 74.05 & 79.48 & 0.15 & -0.02 \\
  & DEMIX & 79.99 & 87.12 & 68.46 & 72.73 & 63.35 & 72.86 & 78.07 & 82.42 & 71.73 & 86.59 & 91.12 & 74.70 & 79.66 & 0.74 & 0.36 \\
 & BCL & 78.97 & 86.52 & \textbf{70.71} & \textbf{74.58} & 66.26 & 74.55 & 81.70 & 84.63 & 71.99 & 85.06 & 90.51 & 75.78 & 80.46 & -0.06 & -0.19 \\
& CLASSIC & 79.89 & 87.05 & 67.30 & 72.11 & 59.84 & 71.08 & 84.02 & 86.22 & 69.83 & 86.93 & 91.25 & 74.63 & 79.59 & 0.44 & 0.25 \\
 & KD & 78.05 & 85.59 & 69.17 & 73.73 & 67.49 & 75.09 & 82.12 & 84.99 & 72.28 & 81.91 & 88.69 & 75.17 & 80.06 & -0.07 & 0.01 \\
 & EWC & \textbf{80.98} & 87.64 & 65.94 & 71.17 & 65.04 & 73.58 & 82.32 & 85.13 & 71.43 & 83.35 & 89.14 & 74.84 & 79.68 & 0.02 & -0.01 \\
& DER++ & 79.00 & 86.46 & 67.20 & 72.16 & 63.96 & 73.54 & 83.22 & 85.61 & 72.58 & 87.10 & 91.47 & 75.51 & 80.30 & 2.36 & 1.53 \\
 & HAT & 76.42 & 85.16 & 60.70 & 68.79 & 47.37 & 65.69 & 72.33 & 79.13 & 69.97 & 74.04 & 85.14 & 66.80 & 75.65 & -0.13 & -0.29 \\
  & HAT-All & 74.94 & 83.93 & 52.08 & 63.94 & 34.16 & 56.07 & 64.71 & 74.43 & 68.14 & 65.54 & 81.44 & 59.93 & 71.33 & 3.23 & 1.83 \\
& HAT-Adapter & 79.29 & 86.70 & 68.25 & 72.87 & 64.84 & 73.67 & 81.44 & 84.56 & 71.61 & 82.37 & 89.27 & 74.63 & 79.78 & -0.23 & -0.18 \\
 & DAS & 80.34 & 87.16 & 69.36 & 74.01 & \textbf{70.93} & \textbf{77.46} & \textbf{85.99} & \textbf{87.70} & 72.80 & \textbf{88.16} & \textbf{92.30} & \textbf{77.93} & \textbf{81.91} & \textbf{-1.09} & \textbf{-0.60} \\
\specialrule{.1em}{.05em}{.05em}
\specialrule{.1em}{.05em}{.05em}
\end{tabular}}
\vspace{-3mm}
\caption{End-task macro-F1 (MF1), accuracy and forgetting rate results 
for all domains \textit{after the continual DAP-training of all domains}, except for CHEMPORT in the PubMed domain, for which we use micro-F1 following~\citep{DBLP:conf/acl/GururanganMSLBD20,dery2021should,beltagy-etal-2019-scibert}. The results are averages of 5 random seeds (the domain training order is as they appear in the first row). Due to space limits, the results for \textit{different domain orders} and the \textit{standard deviations} are reported in Appendix~\ref{ap:order} and Appendix~\ref{ap:std}, respectively). Non-CL baselines have no forgetting. 
} 
\vspace{-6mm}
\label{tab:dapt_result}
\end{table*}

\vspace{-2mm}
\subsection{Results Analysis and Ablation Study}
\label{sec:results}
\vspace{-1mm}
Due to space limits, \textit{\textbf{Implementation Details}} are given in  Appendix~\ref{ap:imp_detail}. Table~\ref{tab:dapt_result} reports the end-task fine-tuning results of all 15 systems on the 6 datasets. We can see that the proposed DAS outperforms all baselines on average and also achieve the best knowledge transfer (negative forgetting rate). 





{\color{black}(1) DAS is slightly better than Pool on average. 
This may be because (a) some domains are quite different (e.g. camera reviews and ACL papers), which results in some negative transfer in Pool. {\color{black}(b) DAS 
can learn with the general and previous domain knowledge protected by soft-masks.}}

(2). DAS achieves both forgetting prevention and knowledge transfer. Those baselines (KD, EWC, DER++) focusing only on forgetting prevention give poorer performance as they sacrifice accuracy to avoid CF. Those baselines (BCL, CLASSIC and DEMIX) perform knowledge transfer achieve better results but still poorer than DAS. DEMIX has very weak transfer. BCL, which can avoid CF while also achieving some transfer, is weaker than NCL. In general, CL baselines are all poorer than DAS as they don’t have methods to encourage knowledge transfer or they have to rely on adapters.

(3). Directly learning the domains within the LM helps DAS achieve better results than adapter and prompt based methods. {\color{black}DAS is better than adapter-based systems (DAP-Adapter, NCL-Adapter and HAT-Adapter) and prompt-based system (DAP-Prompt).} This is because adapters and prompts do not have sufficient trainable parameters,  which are also randomly initialized and can be hard to train. 

{\color{black}(4).} Using the full LM to learn all tasks rather than using sub-networks (of HAT-based methods) 
makes DAS more effective. HAT performs poorly, indicating it is unsuitable for DAP-training as discussed in Sec.~\ref{sec:intro}. 
Even if we use all features (not only the feature from the corresponding sub-network), we still get poor results (HAT-All) as the features used in DAP-training (in an LM sub-network) are different from features used in end-task fine-tuning (features from the whole LM). 

\textbf{Knowledge transfer and forgetting avoidance.} To see how the models fare on CF and knowledge transfer, we compare the forgetting rates (\textbf{forget R.}) \citep{DBLP:conf/cvpr/LiuSLSS20}, $\frac{1}{t-1}\sum_{k=1}^{t-1}A_{k,k} - A_{t,k}$, where $A_{k,k}$ is the end-task accuracy right after its domain $k$ is DAP-trained, and $A_{t,k}$ is the accuracy of the end-task of domain $k$ after DAP-training the last domain $t$. We average over all end-tasks except the last as the last domain has no forgetting. The higher the forgetting rate is, the more forgetting it has. Negative rates indicate positive knowledge transfer. Clearly, DAS has the strongest negative forgetting rate, indicating it does well on both forgetting prevention and knowledge transfer. NCL, NCL-Adapter, DEMIX, EWC, KD and DER++ all suffer from some forgetting. HAT has no forgetting but it cannot learn well. HAT and BCL have no forgetting but are weak in  transfer.

\textbf{Effectiveness of the proxy KL-divergence loss.} We use proxy KL-divergence loss in the \textit{initialization} function (Sec. \ref{sec:initialization}) to compute the importance of units for general knowledge. We are interested in how good the proxy is. We use two kinds of experiments to provide evidences. 

{\color{black}

{\color{black}\textbf{(1) \textit{Comparing with a sample set of }$D_0$.}}
In some cases, the continual DAP-training users may have the data $D_0$ that was used to pre-train the LM. Then we can just sample a subset from $D_0$ to compute the parameter importance to the general knowledge in the LM. However, since we do not have $D_0$ that was used to pre-train RoBERTa, we use the Wiki data~\citep{DBLP:conf/iclr/MerityX0S17} as the sample set of $D_0$. We choose it as it is a general dataset with a wide topic coverage and was used to pre-train an LM, and it has a similar size as our domain data (around 700M). We conducted two experiments using the data: \textbf{(a) DAS (Wiki+MLM)}, which uses MLM as the loss in the initialization stage to compute the importance of units (to identify the general knowledge) just like any other domains in the continual learning part, and \textbf{(b) DAS (Wiki+KL)}, which uses KL-divergence as in the initialization stage just like the proposed proxy method. The results are given in Table~\ref{tab:wiki}.

\begin{table*}[]
\centering
\resizebox{\textwidth}{!}{
\begin{tabular}{c|ccccccccccccccc}
\specialrule{.2em}{.1em}{.1em}
Domain & \multicolumn{2}{c}{Restaurant} & \multicolumn{2}{c}{ACL} & \multicolumn{2}{c}{AI} & \multicolumn{2}{c}{Phone} & PubMed & \multicolumn{2}{c}{Camera} & \multicolumn{2}{c}{Average} & \multicolumn{2}{c}{Forget R.} \\
 Model & MF1 & Acc & MF1 & Acc & MF1 & Acc & MF1 & Acc & MF1 & MF1 & Acc & MF1 & Acc & MF1 & Acc \\
\specialrule{.1em}{.05em}{.05em}
DAS   (wiki+KL) & \textbf{81.25} & \textbf{87.89} & \textbf{70.89} & \textbf{74.87} & 69.68 & 76.86 & 85.98 & \textbf{87.78} & 72.03 & 86.69 & 91.44 & 77.75 & 81.81 & -0.50 & -0.27 \\
DAS (wiki+MLM) & 80.22 & 87.12 & 68.12 & 72.92 & 68.55 & 76.06 & 83.50 & 86.11 & 71.94 & 86.02 & 91.15 & 76.39 & 80.88 & 0.54 & 0.40 \\
DAS & 80.34 & 87.16 & 69.36 & 74.01 & \textbf{70.93} & \textbf{77.46} & \textbf{85.99} & 87.70 & \textbf{72.80} & \textbf{88.16} & \textbf{92.30} & \textbf{77.93} & \textbf{81.91} & \textbf{-1.09} & \textbf{-0.60} \\
\specialrule{.1em}{.05em}{.05em}
\specialrule{.1em}{.05em}{.05em}
\end{tabular}}
\vspace{-3mm}
\caption{
{\color{black}Results for the Wiki dataset as the sample set of $D_0$ - average of 5 random seeds}
} 
\vspace{-4mm}
\label{tab:wiki}
\end{table*}

We can see that DAS (Wiki + KL) performs similarly to DAS but  outperforms DAS (Wiki + MLM). This indicates that the proposed proxy KL-divergence is more effective. MLM actually adapts the LM to the Wikipedia data, which may not be sufficiently representative of the original data used in pre-training the LM. As a result, it ends up identifying the knowledge that is suitable only for the Wikipedia data. In contrast, the proposed proxy KL-divergence leverages the random dropout mask and measures the robustness, which is less related to a specific domain and thus reflects the (general) knowledge in the original LM better. }


{\color{black}
\textbf{(2) \textit{Comparing general knowledge computed from different domain corpora}.} 
Here, we also provide some indirect evidences to show the effectiveness of the proxy method for computing the importance of units to the general knowledge in the LM. 
We conduct a separate non-CL experiment to compare the attention heads' importance score vectors  after applying the proxy using the data from different domains.\footnote{{\color{black}We use attention heads instead of other units because they are arguably the most important component in a Transformer~\citep{michel2019sixteen,voita2019analyzing,mccarley2019structured}.}} For each domain $i$, we compare its importance vector with the importance vector of every other domain, and 
then average the cosine similarities to get the value for domain $i$. 
We get 0.92 for Restaurant, the same 0.91 for ACL, AI, and Phone, 0.89 for PubMed and 0.92 for Camera. We see that different domains give similar importance values, which indirectly shows that our proxy can approximately identify the common general knowledge.

}

\begin{table*}[]
\centering
\resizebox{\textwidth}{!}{
\begin{tabular}{cc|ccccccccccccccc}
\specialrule{.2em}{.1em}{.1em}
\multirow{2}{*}{Category} & Domain & \multicolumn{2}{c}{Restaurant} & \multicolumn{2}{c}{ACL} & \multicolumn{2}{c}{AI} & \multicolumn{2}{c}{Phone} & PubMed & \multicolumn{2}{c}{Camera} & \multicolumn{2}{c}{Average} & \multicolumn{2}{c}{Forget R.} \\
 & Model & MF1 & Acc & MF1 & Acc & MF1 & Acc & MF1 & Acc & MF1 & MF1 & Acc & MF1 & Acc & MF1 & Acc \\
\specialrule{.1em}{.05em}{.05em}
\multirow{4}{*}{Non-CL} & RoBERTa & 79.81 & 87.00 & 66.11 & 71.26 & 60.98 & 71.85 & 83.75 & 86.08 & 72.38 & 78.82 & 87.03 & 73.64 & 79.27 & \multicolumn{2}{c}{---} \\
 & DAP-RoBERTa & 80.84 & \textbf{87.68} & 68.75 & 73.44 & 68.97 & 75.95 & 82.59 & 85.50 & 72.84 & 84.39 & 89.90 & 76.40 & 80.89 & \multicolumn{2}{c}{---} \\
 & DAP-Adapter & 80.19 & 87.14 & 68.87 & 72.92 & 60.55 & 71.38 & 82.71 & 85.35 & 71.68 & 83.62 & 89.23 & 74.60 & 79.62 & \multicolumn{2}{c}{---} \\
 & DAP-Prompt & 79.00 & 86.45 & 66.66 & 71.35 & 61.47 & 72.36 & 84.17 & 86.53 & \textbf{73.09} & 85.52 & 90.38 & 74.98 & 80.03 & \multicolumn{2}{c}{---} \\
\specialrule{.1em}{.05em}{.05em}
& {\color{black}DAS (random)} & {\color{black}79.79} & {\color{black}86.84} & 	{\color{black}68.34} & 	{\color{black}73.02} & 	{\color{black}68.62} & 	{\color{black}76.17} & 	{\color{black}84.92} & 	{\color{black}87.02} & 	{\color{black}72.73} & 	{\color{black}85.92} & 	{\color{black}91.15} & 	{\color{black}76.72} & 	{\color{black}81.15} &  {\color{black}0.45} & 	{\color{black}0.26} \\
\multirow{4}{*}{\multirow{2}{*}{\begin{tabular}[c]{@{}c@{}}CL\\ DAP-train\end{tabular}}}  & DAS (w/o contrast) & 81.06 & 87.55 & \textbf{70.39} & \textbf{74.39} & 67.60 & 75.32 & 83.53 & 86.00 & 72.01 & 84.48 & 90.06 & 76.51 & 80.89 & -0.54 & -0.23 \\
 & DAS (w/o softmask) & 80.48 & 87.27 & 69.92 & 74.39 & 67.73 & 75.78 & 84.00 & 86.37 & 73.03 & 87.96 & 92.08 & 77.19 & 81.48 & -0.24 & -0.12 \\
 & DAS (w/o initialization) & \textbf{81.30}  & 87.79 & 68.35 & 73.10 & 67.82 & 75.73 & 85.13 & 86.98 & 71.82 & 87.25 & 91.57 & 76.95 & 81.16 & 0.70 & 0.48 \\
& DAS (domain-specific) & 80.95 & \textbf{87.68} & 69.18 & 73.21 & 68.92 & 	76.27 & 83.89 & 86.26 & 72.46 & 86.74 & 91.53 & 77.02 & 81.23 & 	-0.07 & 0.08 \\
 & DAS & 80.34 & 87.16 & 69.36 & 74.01 & \textbf{70.93} & \textbf{77.46} & \textbf{85.99} & \textbf{87.70} & 72.80 & \textbf{88.16} & \textbf{92.30} & \textbf{77.93} & \textbf{81.91} & \textbf{-1.09} & \textbf{-0.60} \\
\specialrule{.1em}{.05em}{.05em}
\specialrule{.1em}{.05em}{.05em}
\end{tabular}}
\vspace{-3mm}
\caption{
Ablation results - averages of 5 random seeds. See \textit{standard deviations} in Appendix~\ref{ap:std}. 
} 
\vspace{-6mm}
\label{tab:ablation}
\end{table*}

\textbf{Ablation.} 
We want to know if the proposed (1) initialization (Sec. \ref{sec:initialization}), (2) soft-masking, and (3) contrastive learning are helpful.
To answer (1), we conduct the ablation \textbf{DAS (w/o initialization)}, where we remove the initialization and directly do the continual learning given no consideration to the general knowledge in the LM. To answer (2), we conduct the ablations (1) \textbf{DAS (w/o softmask)}, where we remove the soft-masks, and only use contrastive learning based on Eq.~\ref{eq:contrast} (with the second term in the denominator removed); {\color{black}and (2) \textbf{DAS (random)} with randomly generated importance scores to do soft-masking and contrastive learning.} To answer (3), we conduct two ablations: \textbf{(i) DAS (w/o contrast)} where we remove the contrastive loss and only soft-mask
according to the importance; \textbf{(ii) DAS (domain-specific)} where we contrast domain-specific and learned knowledge (Sec.~\ref{sec:training}).
Table~\ref{tab:ablation} shows that the full DAS is the best on average and for most domains, indicating that every component contributes. Additional observations are: (1) DAS's gain is partially from the preserved general knowledge. We can see DAS (w/o initialization) is poorer on average; (2) Soft-masking helps as DAS (w/o softmask) is 
poorer than DAS. This is reasonable because {\color{black} soft masking} can preserve learned domains. {\color{black}Besides, our gradient-based mask is informative as DAS (random) is worse than DAS}; (3) Contrastive learning is effective as DAS (w/o contrast) and DAS (domain-specific) are both poorer, indicating the contrastive learning in DAS can help {\color{black} learn good representations}

\vspace{-1mm}
\section{Conclusion}
\vspace{-1mm}
This paper proposed a novel method DAS for the continual DAP-training of an LM. {\color{black}It has three key ideas: (1) Preserving the important previous knowledge by soft-masking units according to their importance to overcome CF and to facilitate knowledge transfer.~(2) Using a novel proxy to compute the importance of units to the general knowledge in the LM.} (3) Learning complementary representations for knowledge integration. 
A set of techniques is proposed to achieve them. Extensive experiments showed the effectiveness of DAS. {\color{black}The current approach involves two functions in learning. We will study how to combine them to further improve the results in the future.} 

\section*{Acknowledgements}
The work of Zixuan Ke, Gyuhak Kim, and Bing Liu was supported in part by a research contract from KDDI, a research contract from DARPA (HR001120C0023), and three NSF grants (IIS-1910424, IIS-1838770, and CNS-2225427). 




\bibliography{iclr2023_conference}

\begin{thebibliography}{64}
\providecommand{\natexlab}[1]{#1}
\providecommand{\url}[1]{\texttt{#1}}
\expandafter\ifx\csname urlstyle\endcsname\relax
  \providecommand{\doi}[1]{doi: #1}\else
  \providecommand{\doi}{doi: \begingroup \urlstyle{rm}\Url}\fi

\bibitem[Alsentzer et~al.(2019)Alsentzer, Murphy, Boag, Weng, Jin, Naumann, and
  McDermott]{alsentzer2019publicly}
Emily Alsentzer, John~R Murphy, Willie Boag, Wei-Hung Weng, Di~Jin, Tristan
  Naumann, and Matthew McDermott.
\newblock Publicly available clinical bert embeddings.
\newblock \emph{arXiv preprint arXiv:1904.03323}, 2019.

\bibitem[Beltagy et~al.(2019)Beltagy, Lo, and Cohan]{beltagy-etal-2019-scibert}
Iz~Beltagy, Kyle Lo, and Arman Cohan.
\newblock {S}ci{BERT}: A pretrained language model for scientific text.
\newblock In \emph{EMNLP-IJCNLP}, 2019.

\bibitem[Brix et~al.(2020)Brix, Bahar, and Ney]{brix-etal-2020-successfully}
Christopher Brix, Parnia Bahar, and Hermann Ney.
\newblock Successfully applying the stabilized lottery ticket hypothesis to the
  transformer architecture.
\newblock In \emph{ACL}, July 2020.

\bibitem[Buzzega et~al.(2020)Buzzega, Boschini, Porrello, Abati, and
  Calderara]{buzzega2020dark}
Pietro Buzzega, Matteo Boschini, Angelo Porrello, Davide Abati, and Simone
  Calderara.
\newblock Dark experience for general continual learning: a strong, simple
  baseline.
\newblock \emph{arXiv preprint arXiv:2004.07211}, 2020.

\bibitem[Chakrabarty et~al.(2019)Chakrabarty, Hidey, and
  McKeown]{chakrabarty2019imho}
Tuhin Chakrabarty, Christopher Hidey, and Kathleen McKeown.
\newblock Imho fine-tuning improves claim detection.
\newblock \emph{arXiv preprint arXiv:1905.07000}, 2019.

\bibitem[Chen et~al.(2020)Chen, Kornblith, Norouzi, and Hinton]{chen2020simple}
Ting Chen, Simon Kornblith, Mohammad Norouzi, and Geoffrey Hinton.
\newblock A simple framework for contrastive learning of visual
  representations.
\newblock In \emph{International conference on machine learning}, pp.\
  1597--1607. PMLR, 2020.

\bibitem[Chuang et~al.(2020)Chuang, Su, and Chen]{chuang2020lifelong}
Yung-Sung Chuang, Shang-Yu Su, and Yun-Nung Chen.
\newblock Lifelong language knowledge distillation.
\newblock \emph{arXiv preprint arXiv:2010.02123}, 2020.

\bibitem[Dery et~al.(2021)Dery, Michel, Talwalkar, and Neubig]{dery2021should}
Lucio~M Dery, Paul Michel, Ameet Talwalkar, and Graham Neubig.
\newblock Should we be pre-training? an argument for end-task aware training as
  an alternative.
\newblock \emph{arXiv preprint arXiv:2109.07437}, 2021.

\bibitem[Devlin et~al.(2019)Devlin, Chang, Lee, and
  Toutanova]{DBLP:conf/naacl/DevlinCLT19}
Jacob Devlin, Ming{-}Wei Chang, Kenton Lee, and Kristina Toutanova.
\newblock {BERT:} pre-training of deep bidirectional transformers for language
  understanding.
\newblock In Jill Burstein, Christy Doran, and Thamar Solorio (eds.),
  \emph{NAACL-HLT}, 2019.

\bibitem[Ding et~al.(2008)Ding, Liu, and Yu]{ding2008holistic}
Xiaowen Ding, Bing Liu, and Philip~S Yu.
\newblock A holistic lexicon-based approach to opinion mining.
\newblock In \emph{Proceedings of the 2008 international conference on web
  search and data mining}, 2008.

\bibitem[Greco et~al.(2019)Greco, Plank, Fern{\'a}ndez, and
  Bernardi]{greco2019psycholinguistics}
Claudio Greco, Barbara Plank, Raquel Fern{\'a}ndez, and Raffaella Bernardi.
\newblock Psycholinguistics meets continual learning: Measuring catastrophic
  forgetting in visual question answering.
\newblock \emph{arXiv preprint arXiv:1906.04229}, 2019.

\bibitem[Gu et~al.(2021)Gu, Han, Liu, and Huang]{gu2021ppt}
Yuxian Gu, Xu~Han, Zhiyuan Liu, and Minlie Huang.
\newblock Ppt: Pre-trained prompt tuning for few-shot learning.
\newblock \emph{arXiv preprint arXiv:2109.04332}, 2021.

\bibitem[Guo et~al.(2016)Guo, Yao, and Chen]{guo2016dynamic}
Yiwen Guo, Anbang Yao, and Yurong Chen.
\newblock Dynamic network surgery for efficient dnns.
\newblock \emph{NIPS}, 29, 2016.

\bibitem[Gupta et~al.(2020)Gupta, Chaudhary, Runkler, and
  Schuetze]{gupta2020neural}
Pankaj Gupta, Yatin Chaudhary, Thomas Runkler, and Hinrich Schuetze.
\newblock Neural topic modeling with continual lifelong learning.
\newblock In \emph{International Conference on Machine Learning}, pp.\
  3907--3917. PMLR, 2020.

\bibitem[Gururangan et~al.(2020)Gururangan, Marasovic, Swayamdipta, Lo,
  Beltagy, Downey, and Smith]{DBLP:conf/acl/GururanganMSLBD20}
Suchin Gururangan, Ana Marasovic, Swabha Swayamdipta, Kyle Lo, Iz~Beltagy, Doug
  Downey, and Noah~A. Smith.
\newblock Don't stop pretraining: Adapt language models to domains and tasks.
\newblock In \emph{ACL}, 2020.

\bibitem[Gururangan et~al.(2021)Gururangan, Lewis, Holtzman, Smith, and
  Zettlemoyer]{gururangan2021demix}
Suchin Gururangan, Mike Lewis, Ari Holtzman, Noah~A Smith, and Luke
  Zettlemoyer.
\newblock Demix layers: Disentangling domains for modular language modeling.
\newblock \emph{arXiv preprint arXiv:2108.05036}, 2021.

\bibitem[Han et~al.(2015)Han, Pool, Tran, and Dally]{han2015learning}
Song Han, Jeff Pool, John Tran, and William Dally.
\newblock Learning both weights and connections for efficient neural network.
\newblock \emph{Advances in neural information processing systems}, 28, 2015.

\bibitem[He et~al.(2020)He, Fan, Wu, Xie, and Girshick]{he2020momentum}
Kaiming He, Haoqi Fan, Yuxin Wu, Saining Xie, and Ross Girshick.
\newblock Momentum contrast for unsupervised visual representation learning.
\newblock In \emph{CVPR}, pp.\  9729--9738, 2020.

\bibitem[He \& Jaeger(2018)He and Jaeger]{He2018overcoming}
Xu~He and Herbert Jaeger.
\newblock Overcoming catastrophic interference using conceptor-aided
  backpropagation.
\newblock In \emph{ICLR}, 2018.

\bibitem[Hinton et~al.(2015)Hinton, Vinyals, Dean,
  et~al.]{hinton2015distilling}
Geoffrey Hinton, Oriol Vinyals, Jeff Dean, et~al.
\newblock Distilling the knowledge in a neural network.
\newblock \emph{arXiv preprint arXiv:1503.02531}, 2\penalty0 (7), 2015.

\bibitem[Houlsby et~al.(2019)Houlsby, Giurgiu, Jastrzebski, Morrone,
  de~Laroussilhe, Gesmundo, Attariyan, and Gelly]{Houlsby2019Parameter}
Neil Houlsby, Andrei Giurgiu, Stanislaw Jastrzebski, Bruna Morrone, Quentin
  de~Laroussilhe, Andrea Gesmundo, Mona Attariyan, and Sylvain Gelly.
\newblock Parameter-efficient transfer learning for {NLP}.
\newblock In Kamalika Chaudhuri and Ruslan Salakhutdinov (eds.), \emph{ICML},
  2019.

\bibitem[Hu \& Liu(2004)Hu and Liu]{hu2004mining}
Minqing Hu and Bing Liu.
\newblock Mining and summarizing customer reviews.
\newblock In \emph{Proceedings of ACM SIGKDD}, 2004.

\bibitem[Huang et~al.(2021)Huang, Zhang, Chen, Wang, and
  Yang]{huang2021continual}
Yufan Huang, Yanzhe Zhang, Jiaao Chen, Xuezhi Wang, and Diyi Yang.
\newblock Continual learning for text classification with information
  disentanglement based regularization.
\newblock \emph{arXiv preprint arXiv:2104.05489}, 2021.

\bibitem[Jang et~al.(2021)Jang, Ye, Yang, Shin, Han, Kim, Choi, and
  Seo]{jang2021continual}
Joel Jang, Seonghyeon Ye, Sohee Yang, Joongbo Shin, Janghoon Han, Gyeonghun
  Kim, Stanley~Jungkyu Choi, and Minjoon Seo.
\newblock Towards continual knowledge learning of language models, 2021.

\bibitem[Jin et~al.(2021)Jin, Zhang, Zhu, Xiao, Li, Wei, Arnold, and
  Ren]{jin2021lifelong}
Xisen Jin, Dejiao Zhang, Henghui Zhu, Wei Xiao, Shang-Wen Li, Xiaokai Wei,
  Andrew Arnold, and Xiang Ren.
\newblock Lifelong pretraining: Continually adapting language models to
  emerging corpora.
\newblock \emph{arXiv preprint arXiv:2110.08534}, 2021.

\bibitem[Jurgens et~al.(2018)Jurgens, Kumar, Hoover, McFarland, and
  Jurafsky]{DBLP:journals/tacl/JurgensKHMJ18}
David Jurgens, Srijan Kumar, Raine Hoover, Daniel~A. McFarland, and Dan
  Jurafsky.
\newblock Measuring the evolution of a scientific field through citation
  frames.
\newblock \emph{TACL}, 2018.

\bibitem[Ke et~al.(2021{\natexlab{a}})Ke, Liu, Ma, Xu, and
  Shu]{ke2021achieving}
Zixuan Ke, Bing Liu, Nianzu Ma, Hu~Xu, and Lei Shu.
\newblock Achieving forgetting prevention and knowledge transfer in continual
  learning.
\newblock \emph{Advances in Neural Information Processing Systems}, 34,
  2021{\natexlab{a}}.

\bibitem[Ke et~al.(2021{\natexlab{b}})Ke, Liu, Xu, and Shu]{ke2021Classic}
Zixuan Ke, Bing Liu, Hu~Xu, and Lei Shu.
\newblock Classic: Continual and contrastive learning of aspect sentiment
  classification tasks.
\newblock In \emph{EMNLP}, 2021{\natexlab{b}}.

\bibitem[Ke et~al.(2021{\natexlab{c}})Ke, Xu, and Liu]{ke2021adapting}
Zixuan Ke, Hu~Xu, and Bing Liu.
\newblock Adapting bert for continual learning of a sequence of aspect
  sentiment classification tasks.
\newblock In \emph{NAACL}, pp.\  4746--4755, 2021{\natexlab{c}}.

\bibitem[Ke et~al.(2022{\natexlab{a}})Ke, Lin, Shao, Xu, Shu, and
  Liu]{ke2022cpt}
Zixuan Ke, Haowei Lin, Yijia Shao, Hu~Xu, Lei Shu, and Bing Liu.
\newblock Continual training of language models for few-shot learning.
\newblock In \emph{Empirical Methods in Natural Language Processing (EMNLP)},
  2022{\natexlab{a}}.

\bibitem[Ke et~al.(2022{\natexlab{b}})Ke, Shao, Lin, Xu, Shu, and
  Liu]{ke2022dga}
Zixuan Ke, Yijia Shao, Haowei Lin, Hu~Xu, Lei Shu, and Bing Liu.
\newblock Adapting a language model while preserving its general knowledge.
\newblock In \emph{Empirical Methods in Natural Language Processing (EMNLP)},
  2022{\natexlab{b}}.

\bibitem[Kim et~al.(2022)Kim, Xiao, Konishi, Ke, and Liu]{kim2022theoretical}
Gyuhak Kim, Changnan Xiao, Tatsuya Konishi, Zixuan Ke, and Bing Liu.
\newblock A theoretical study on solving continual learning.
\newblock In \emph{Advances in Neural Information Processing Systems}, 2022.

\bibitem[Kirkpatrick et~al.(2016)Kirkpatrick, Pascanu, Rabinowitz, Veness,
  Desjardins, Rusu, Milan, Quan, Ramalho, Grabska{-}Barwinska, Hassabis,
  Clopath, Kumaran, and Hadsell]{Kirkpatrick2017overcoming}
James Kirkpatrick, Razvan Pascanu, Neil~C. Rabinowitz, Joel Veness, Guillaume
  Desjardins, Andrei~A. Rusu, Kieran Milan, John Quan, Tiago Ramalho, Agnieszka
  Grabska{-}Barwinska, Demis Hassabis, Claudia Clopath, Dharshan Kumaran, and
  Raia Hadsell.
\newblock Overcoming catastrophic forgetting in neural networks.
\newblock \emph{CoRR}, 2016.

\bibitem[Kringelum et~al.(2016)Kringelum, Kjaerulff, Brunak, Lund, Oprea, and
  Taboureau]{kringelum2016chemprot}
Jens Kringelum, Sonny~Kim Kjaerulff, S{\o}ren Brunak, Ole Lund, Tudor~I Oprea,
  and Olivier Taboureau.
\newblock Chemprot-3.0: a global chemical biology diseases mapping.
\newblock \emph{Database}, 2016, 2016.

\bibitem[Lai et~al.(2021)Lai, Zhang, Liu, Chang, Liao, Chuang, Qian, Khurana,
  Cox, and Glass]{lai2021parp}
Cheng-I~Jeff Lai, Yang Zhang, Alexander~H Liu, Shiyu Chang, Yi-Lun Liao,
  Yung-Sung Chuang, Kaizhi Qian, Sameer Khurana, David Cox, and Jim Glass.
\newblock Parp: Prune, adjust and re-prune for self-supervised speech
  recognition.
\newblock \emph{NeurIPS}, 34, 2021.

\bibitem[Lange et~al.(2019)Lange, Aljundi, Masana, and
  Tuytelaars]{DBLP:journals/corr/abs-1909-08383}
Matthias~De Lange, Rahaf Aljundi, Marc Masana, and Tinne Tuytelaars.
\newblock Continual learning: {A} comparative study on how to defy forgetting
  in classification tasks.
\newblock \emph{CoRR}, 2019.

\bibitem[Lee et~al.(2020)Lee, Yoon, Kim, Kim, Kim, So, and
  Kang]{lee2020biobert}
Jinhyuk Lee, Wonjin Yoon, Sungdong Kim, Donghyeon Kim, Sunkyu Kim, Chan~Ho So,
  and Jaewoo Kang.
\newblock Biobert: a pre-trained biomedical language representation model for
  biomedical text mining.
\newblock \emph{Bioinformatics}, 36\penalty0 (4):\penalty0 1234--1240, 2020.

\bibitem[Lester et~al.(2021)Lester, Al{-}Rfou, and
  Constant]{DBLP:conf/emnlp/LesterAC21}
Brian Lester, Rami Al{-}Rfou, and Noah Constant.
\newblock The power of scale for parameter-efficient prompt tuning.
\newblock In Marie{-}Francine Moens, Xuanjing Huang, Lucia Specia, and
  Scott~Wen{-}tau Yih (eds.), \emph{EMNLP}, 2021.

\bibitem[Li et~al.(2021)Li, Cotterell, and Sachan]{li2021differentiable}
Jiaoda Li, Ryan Cotterell, and Mrinmaya Sachan.
\newblock Differentiable subset pruning of transformer heads.
\newblock \emph{Transactions of the Association for Computational Linguistics},
  9:\penalty0 1442--1459, 2021.

\bibitem[Li et~al.(2019)Li, Zhao, Church, and Elhoseiny]{li2019compositional}
Yuanpeng Li, Liang Zhao, Kenneth Church, and Mohamed Elhoseiny.
\newblock Compositional language continual learning.
\newblock In \emph{International Conference on Learning Representations}, 2019.

\bibitem[Liu et~al.(2020)Liu, Su, Liu, Schiele, and
  Sun]{DBLP:conf/cvpr/LiuSLSS20}
Yaoyao Liu, Yuting Su, An{-}An Liu, Bernt Schiele, and Qianru Sun.
\newblock Mnemonics training: Multi-class incremental learning without
  forgetting.
\newblock In \emph{CVPR}, 2020.

\bibitem[Liu et~al.(2019)Liu, Ott, Goyal, Du, Joshi, Chen, Levy, Lewis,
  Zettlemoyer, and Stoyanov]{DBLP:journals/corr/abs-1907-11692}
Yinhan Liu, Myle Ott, Naman Goyal, Jingfei Du, Mandar Joshi, Danqi Chen, Omer
  Levy, Mike Lewis, Luke Zettlemoyer, and Veselin Stoyanov.
\newblock Roberta: {A} robustly optimized {BERT} pretraining approach.
\newblock \emph{CoRR}, 2019.

\bibitem[Lo et~al.(2020)Lo, Wang, Neumann, Kinney, and
  Weld]{DBLP:conf/acl/LoWNKW20}
Kyle Lo, Lucy~Lu Wang, Mark Neumann, Rodney Kinney, and Daniel~S. Weld.
\newblock {S2ORC:} the semantic scholar open research corpus.
\newblock In Dan Jurafsky, Joyce Chai, Natalie Schluter, and Joel~R. Tetreault
  (eds.), \emph{ACL}, 2020.

\bibitem[Luan et~al.(2018)Luan, He, Ostendorf, and
  Hajishirzi]{DBLP:conf/emnlp/LuanHOH18}
Yi~Luan, Luheng He, Mari Ostendorf, and Hannaneh Hajishirzi.
\newblock Multi-task identification of entities, relations, and coreference for
  scientific knowledge graph construction.
\newblock In Ellen Riloff, David Chiang, Julia Hockenmaier, and Jun'ichi Tsujii
  (eds.), \emph{ACL}, 2018.

\bibitem[Madotto et~al.(2020)Madotto, Lin, Zhou, Moon, Crook, Liu, Yu, Cho, and
  Wang]{madotto2020continual}
Andrea Madotto, Zhaojiang Lin, Zhenpeng Zhou, Seungwhan Moon, Paul Crook, Bing
  Liu, Zhou Yu, Eunjoon Cho, and Zhiguang Wang.
\newblock Continual learning in task-oriented dialogue systems.
\newblock \emph{arXiv preprint arXiv:2012.15504}, 2020.

\bibitem[McCarley et~al.(2019)McCarley, Chakravarti, and
  Sil]{mccarley2019structured}
JS~McCarley, Rishav Chakravarti, and Avirup Sil.
\newblock Structured pruning of a bert-based question answering model.
\newblock \emph{arXiv preprint arXiv:1910.06360}, 2019.

\bibitem[Merity et~al.(2017)Merity, Xiong, Bradbury, and
  Socher]{DBLP:conf/iclr/MerityX0S17}
Stephen Merity, Caiming Xiong, James Bradbury, and Richard Socher.
\newblock Pointer sentinel mixture models.
\newblock In \emph{5th International Conference on Learning Representations,
  {ICLR} 2017, Toulon, France, April 24-26, 2017, Conference Track
  Proceedings}. OpenReview.net, 2017.

\bibitem[Michel et~al.(2019)Michel, Levy, and Neubig]{michel2019sixteen}
Paul Michel, Omer Levy, and Graham Neubig.
\newblock Are sixteen heads really better than one?
\newblock \emph{Advances in neural information processing systems}, 32, 2019.

\bibitem[Mitchell et~al.(2022)Mitchell, Lin, Bosselut, Finn, and
  Manning]{DBLP:conf/iclr/MitchellLBFM22}
Eric Mitchell, Charles Lin, Antoine Bosselut, Chelsea Finn, and Christopher~D.
  Manning.
\newblock Fast model editing at scale.
\newblock In \emph{The Tenth International Conference on Learning
  Representations, {ICLR} 2022, Virtual Event, April 25-29, 2022}.
  OpenReview.net, 2022.

\bibitem[Ni et~al.(2019)Ni, Li, and McAuley]{DBLP:conf/emnlp/NiLM19}
Jianmo Ni, Jiacheng Li, and Julian~J. McAuley.
\newblock Justifying recommendations using distantly-labeled reviews and
  fine-grained aspects.
\newblock In Kentaro Inui, Jing Jiang, Vincent Ng, and Xiaojun Wan (eds.),
  \emph{EMNLP}, pp.\  188--197. Association for Computational Linguistics,
  2019.

\bibitem[Pfeiffer et~al.(2020)Pfeiffer, Kamath, R{\"u}ckl{\'e}, Cho, and
  Gurevych]{pfeiffer2020adapterfusion}
Jonas Pfeiffer, Aishwarya Kamath, Andreas R{\"u}ckl{\'e}, Kyunghyun Cho, and
  Iryna Gurevych.
\newblock Adapterfusion: Non-destructive task composition for transfer
  learning.
\newblock In \emph{EACL}, 2020.

\bibitem[Qin et~al.(2022)Qin, Zhang, Lin, Liu, Li, Sun, and
  Zhou]{DBLP:conf/acl/QinZLL0SZ22}
Yujia Qin, Jiajie Zhang, Yankai Lin, Zhiyuan Liu, Peng Li, Maosong Sun, and Jie
  Zhou.
\newblock {ELLE:} efficient lifelong pre-training for emerging data.
\newblock In Smaranda Muresan, Preslav Nakov, and Aline Villavicencio (eds.),
  \emph{Findings of the Association for Computational Linguistics: {ACL} 2022,
  Dublin, Ireland, May 22-27, 2022}, pp.\  2789--2810. Association for
  Computational Linguistics, 2022.
\newblock URL \url{https://aclanthology.org/2022.findings-acl.220}.

\bibitem[Rebuffi et~al.(2017)Rebuffi, Kolesnikov, Sperl, and
  Lampert]{Rebuffi2017}
Sylvestre{-}Alvise Rebuffi, Alexander Kolesnikov, Georg Sperl, and Christoph~H.
  Lampert.
\newblock icarl: Incremental classifier and representation learning.
\newblock In \emph{CVPR}, 2017.

\bibitem[Seff et~al.(2017)Seff, Beatson, Suo, and Liu]{Seff2017continual}
Ari Seff, Alex Beatson, Daniel Suo, and Han Liu.
\newblock Continual learning in generative adversarial nets.
\newblock \emph{CoRR}, abs/1705.08395, 2017.

\bibitem[Serr{\`{a}} et~al.(2018)Serr{\`{a}}, Suris, Miron, and
  Karatzoglou]{Serra2018overcoming}
Joan Serr{\`{a}}, Didac Suris, Marius Miron, and Alexandros Karatzoglou.
\newblock Overcoming catastrophic forgetting with hard attention to the task.
\newblock In \emph{ICML}, 2018.

\bibitem[Shen et~al.(2019)Shen, Zeng, and Jin]{shen-etal-2019-progressive}
Yilin Shen, Xiangyu Zeng, and Hongxia Jin.
\newblock A progressive model to enable continual learning for semantic slot
  filling.
\newblock In \emph{EMNLP-IJCNLP}, November 2019.

\bibitem[Shin et~al.(2017)Shin, Lee, Kim, and Kim]{Shin2017continual}
Hanul Shin, Jung~Kwon Lee, Jaehong Kim, and Jiwon Kim.
\newblock Continual learning with deep generative replay.
\newblock In \emph{NIPS}, 2017.

\bibitem[Sun et~al.(2019)Sun, Qiu, Xu, and Huang]{sun2019fine}
Chi Sun, Xipeng Qiu, Yige Xu, and Xuanjing Huang.
\newblock How to fine-tune bert for text classification?
\newblock In \emph{China national conference on Chinese computational
  linguistics}, pp.\  194--206. Springer, 2019.

\bibitem[Sun et~al.(2020)Sun, Ho, and Lee]{sun2020lamol}
Fan-Keng Sun, Cheng-Hao Ho, and Hung-Yi Lee.
\newblock Lamol: Language modeling is all you need for lifelong language
  learning.
\newblock In \emph{ICLR}, 2020.
\newblock URL \url{https://openreview.net/forum?id=Skgxcn4YDS}.

\bibitem[Voita et~al.(2019)Voita, Talbot, Moiseev, Sennrich, and
  Titov]{voita2019analyzing}
Elena Voita, David Talbot, Fedor Moiseev, Rico Sennrich, and Ivan Titov.
\newblock Analyzing multi-head self-attention: Specialized heads do the heavy
  lifting, the rest can be pruned.
\newblock \emph{arXiv preprint arXiv:1905.09418}, 2019.

\bibitem[Wang et~al.(2020{\natexlab{a}})Wang, Tang, Duan, Wei, Huang, Cao,
  Jiang, Zhou, et~al.]{wang2020k}
Ruize Wang, Duyu Tang, Nan Duan, Zhongyu Wei, Xuanjing Huang, Guihong Cao,
  Daxin Jiang, Ming Zhou, et~al.
\newblock K-adapter: Infusing knowledge into pre-trained models with adapters.
\newblock \emph{arXiv preprint arXiv:2002.01808}, 2020{\natexlab{a}}.

\bibitem[Wang et~al.(2020{\natexlab{b}})Wang, Mehta, P{\'o}czos, and
  Carbonell]{wang2020efficient}
Zirui Wang, Sanket~Vaibhav Mehta, Barnab{\'a}s P{\'o}czos, and Jaime Carbonell.
\newblock Efficient meta lifelong-learning with limited memory.
\newblock In \emph{EMNLP}, 2020{\natexlab{b}}.

\bibitem[Wortsman et~al.(2020)Wortsman, Ramanujan, Liu, Kembhavi, Rastegari,
  Yosinski, and Farhadi]{wortsman2020supermasks}
Mitchell Wortsman, Vivek Ramanujan, Rosanne Liu, Aniruddha Kembhavi, Mohammad
  Rastegari, Jason Yosinski, and Ali Farhadi.
\newblock Supermasks in superposition.
\newblock In H.~Larochelle, M.~Ranzato, R.~Hadsell, M.~F. Balcan, and H.~Lin
  (eds.), \emph{NeurIPS}, 2020.

\bibitem[Xu et~al.(2019)Xu, Liu, Shu, and Yu]{DBLP:conf/naacl/XuLSY19}
Hu~Xu, Bing Liu, Lei Shu, and Philip~S. Yu.
\newblock {BERT} post-training for review reading comprehension and
  aspect-based sentiment analysis.
\newblock In Jill Burstein, Christy Doran, and Thamar Solorio (eds.),
  \emph{NAACL-HLT}, 2019.

\end{thebibliography}
\bibliographystyle{iclr2023_conference}

\appendix
\null\newpage








\section{Datasets Details}
\label{ap:dataset}
Table \ref{tab:dataset} in the main paper has already showed the number of examples in each dataset. Here we provide additional details about the 4 types of end-tasks.

(1) \textbf{(Phone, Camera and Restaurant) Aspect Sentiment Classification (ASC)} is defined as follows: given an aspect or product feature (e.g., \textit{picture quality} in a camera review) and a review sentence containing the aspect in a domain or product category (e.g., camera), classify if the sentence expresses a positive, negative, or neutral (no opinion) sentiment or polarity about the aspect (for the Phone and Camera datasets, there are only negative and positive polarities in the data).

(2) \textbf{(ACL) Citation intent classification} is defined as follows: given a citing sentence (a sentence contains a citation), classify if the sentence expresses a citation function among ``background'', ``motivation'', ``uses'', ``extension'' and ``comparison or contrast future''.

(3) \textbf{(AI) Relation classification} is defined as follows: given a within-sentence word sequence span containing a pair of entities, classify if the span expresses a relation among ``feature of'', ``conjunction'', ``evaluate for'', ``hyponym of'', ``used for'', ``part of'' and ``compare''.

(4) \textbf{(PubMed) Chemical-protein interaction classification} is defined as follows: given a span containing a pair of chemical and protein, classify if the span expresses a chemical-protein interaction among ``downregulator'', ``substrate'', ``indirect-upregulator'', ``indirect-downregulator'', ``agnonist'', ``activator'', ``product of'', ``agonist-activator'', ``inhibitor'', ``upregulator'', ``substrate product of'', ``agonist-inhibitor''and ``antagonist''.

\section{Baseline Details}
\label{ap:baseline}

\textbf{Non-Continual Learning Baselines}: Each of these baselines builds a separate model for each
task independently.
It thus has no knowledge transfer or CF. 

(1) \textbf{Non-DAP-training (RoBERTa)}~\cite{DBLP:journals/corr/abs-1907-11692} uses the original RoBERTa for the end-task fine-tuning without any DAP-training. This is the only one without any DAP-training. All the following baselines use the masked language model loss (MLM) for DAP-training. 

(2) \textbf{DAP-training using masked language model loss (DAP-RoBERTa)} is the existing DAP-training method in~\cite{DBLP:conf/acl/GururanganMSLBD20}. To our knowledge, the existing DAP-training systems are all based on the MLM loss. 

(3) \textbf{DAP-training using adapter-tuning }~\cite{madotto2020continual,Houlsby2019Parameter} adds small adapter layers between layers of Transformer for DAP-training. We follow the adapter design in~\cite{madotto2020continual,Houlsby2019Parameter}: An adapter is simply a 2 layers of fully connected network. During DAP-training, the Transformer is fixed, only the added adapters are trainable. The bottleneck size (adapter size) is set to 128. 
During end-task fine-tuning, both RoBERTa and the adapters are trainable to ensure fair comparison.

(4) \textbf{DAP-training using prompt-tuning }~\cite{DBLP:conf/emnlp/LesterAC21} adds a sequence of real vector tokens (called virtual tokens or prompt tokens) to the end of the original sequence. In DAP-training, RoBERTa (the LM) is fixed and only the prompt tokens are trained. In end-task fine-tuning, both LM and the trained prompt are trainable. We initialize 100 tokens and set the learning rate of the prompt token to 0.3 in DAP-training, following the setting in \cite{DBLP:conf/emnlp/LesterAC21}.

\vspace{+2mm}
\noindent
\textbf{Continual Learning (CL) Baselines.}

(5) \textbf{Naive continual learning (NCL)} is a naive extension of \cite{DBLP:conf/acl/GururanganMSLBD20}, which continually/incrementally DAP-trains the LM to learn all domains using MLM loss with no mechanism to deal with CF.

(6) \textbf{Continual learning with adapter (NCL-Adapter)}~\cite{Houlsby2019Parameter} is similar to the adapter based system. The only difference is that the same set of adapters is shared across all domains, rather than using a new adapter for each new domain.

(7) \textbf{DEMIX (DEMIX)}~\cite{gururangan2021demix} is a recent model to adapt pre-trained LM with new domains. It adds a new adapter once a new domain arrives (network expansion is needed) and initializes the new adapter with the parameters of the previous trained adapter nearest to the new domain data. They use the perplexity on a held-out sample to choose the most probable adapter. For fair comparison, we use the same size as $\{\bm{x}^{\text{sub}}_n\}$ as the held-out samples.

(8) \textbf{Hard attention to overcome forgetting (HAT-Adapter)}~\cite{ke2021adapting} is derived from HAT~\cite{Serra2018overcoming}, the state-of-the-art parameter-isolation based method with almost no forgetting. However, HAT requires task id information in end-task fine-tuning (DAS works in \textit{domain-agnostic} manner and does not need the task id information; see Sec.~\ref{sec:intro}). HAT also needs to train an addition task embedding to mask each layer of the network which makes the DAP-training inefficient. 

(9) \textbf{Continual learning plugin with capsule(BCL)}~\cite{ke2021adapting} is a continual learning model that can avoid forgetting and encourage knowledge transfer. It is similar to NCL-Adapter. The difference is that its adapters consist of two modules, one is a capsule network (a new capsule is added once a new domain arrives) to encourage transfer and the other one is similar to HAT to avoid forgetting. Similar to HAT, task/domain information is needed in end-task fine-tuning. We replace the backbone network from BERT with RoBERTa for fair comparison. 

(10) \textbf{Continual learning plugin with contrastive transfer (CLASSIC)}~\cite{ke2021Classic} is a continual learning model that can avoid forgetting and encourage knowledge transfer via contrasting loss. It is similar to HAT. but 3 additional contrastive loss are used for distillation, knowledge transfer and supervised contrast. Since DAS is working on unsupervised data, we remove the supervised contrastive loss. Similar to HAT, task information is needed in end-task fine-tuning. We replace the backbone network from BERT with RoBERTa for fair comparison. 

(11) \textbf{Knowledge distillation (KD)}~\cite{hinton2015distilling}
minimizes the representational deviation between the learned representation and the new representation in DAP-training. We
compute the KL divergence between the representations (the output before the masked language model prediction head) of each token of the previous DAP-trained LM and current LM as the distillation loss.

(12) \textbf{EWC}~\cite{buzzega2020dark} is a popular
regularization-based method which adopts elastic weights consolidation to add $L_2$ regularization to parameter changes.

(13) \textbf{DER++}~\cite{buzzega2020dark} is a recent replay method using distillation to regularize the new task training. We store 16.4K tokens for each learned domain as the memory, which is the largest memory we can use for the system to run.

(14) \textbf{HAT}~\cite{Serra2018overcoming} is used in the Transformer layers (including self-attention, intermediate and output layers) rather than the added adapter layers. Additional task embedding and task information for end-task fine-tuning are needed.




\section{Implementation Details}
\label{ap:imp_detail}

\textbf{Architecture.} We adopt $\text{RoBERTa}_{\textbf{BASE}}$ as our backbone LM. 
A masked language model head is applied for DAP-training. The end-task fine-tuning of RoBERTa follows the standard practice. 
For the three ASC tasks (see Table~\ref{tab:dataset}), we adopt the ASC formulation in \cite{DBLP:conf/naacl/XuLSY19}, where the aspect (e.g., ``\textit{sound}'') and review sentence (e.g., ``\textit{The sound is great}'') are concatenated via \texttt{</s>}. 

\textbf{Hyperparameters.} 
Unless otherwise stated, the same hyper-parameters are used in all experiments. The maximum input length is set to 164 which is sufficient for all datasets. Adam optimizer is used for both DAP-training and end-task fine-tuning. The max sequence length is also set to 164. 

\textbf{DAP-training.} The learning rate is set to 1e-4 and batch size to 256. We train 2.5K steps for each domain, roughly a full pass through the domain data, following \cite{DBLP:conf/acl/GururanganMSLBD20,DBLP:conf/naacl/XuLSY19}. The subset of data $\{\bm{x}^{\text{sub}}_n\}$ for computing $\mathcal{L}^{\text{impt}}$ to determine head importance in Secs.~\ref{sec:initialization} and \ref{sec:after_training} is set to 1.64 Million tokens, 
which is sufficient in our experiments. $\lambda$ in Eq.~\ref{eq:final_loss} is set to 1 and $\tau$ in Eq.~\ref{eq:contrast} is set to 0.05. 

\textbf{End-task fine-tuning.} The learning rate is set to 1e-5 and batch size to 16. We train on end-task fine-tuning datasets for 5 epochs for Restaurant; 10 epochs for ACL, AI and PubMed; and 15 epochs for Phone and Camera. We simply take the results for the last epoch, assuming no validation sets. We empirically found that the above number of epochs gives us stable and convergence results.

\section{DAP-training in Different Orders}
\label{ap:order}

Table~\ref{tab:dapt_result} in the main paper reported the results for the order \texttt{Restaurant} $\to$ \texttt{ACL} $\to$ \texttt{AI} $\to$ \texttt{Phone} $\to$ \texttt{PubMed} $\to$ \texttt{Camera}. We now look at how the order affects the results. Due to the computation intensive nature of DAP-training, we only report the best baseline (NCL) and DAS results with different domain orders. Table~\ref{tab:order} shows NCL and DAS's results of 5 different orders. We can see DAS is always better than NCL, demonstrating the effectiveness of DAS.

\begin{table*}[]
\centering
\resizebox{0.9\textwidth}{!}{
\begin{tabular}{c|cccccc}
\specialrule{.2em}{.1em}{.1em}
\multirow{2}{*}{Random Sequence Order} & \multicolumn{2}{c}{NCL} & \multicolumn{2}{c}{{\color{black}KD}} &  \multicolumn{2}{c}{DAS} \\
 & MF1 & Acc & MF1 & Acc & MF1 & Acc \\
\specialrule{.1em}{.05em}{.05em}
\texttt{Restaurant} $\to$ \texttt{ACL} $\to$ \texttt{AI} $\to$ \texttt{Phone} $\to$ \texttt{PubMed} $\to$ \texttt{Camera} & 76.36 & 80.77  & {\color{black}75.17} & {\color{black}80.06} & \textbf{77.93} & \textbf{81.91} \\
\texttt{Phone} $\to$ \texttt{AI} $\to$ \texttt{PubMed} $\to$ \texttt{Camera} $\to$ \texttt{Restaurant} $\to$ \texttt{ACL}  & 76.06 & 80.69 & {\color{black}75.49}& {\color{black}80.39} & \textbf{76.90} & \textbf{81.10} \\
\texttt{PubMed} $\to$ \texttt{Camera} $\to$ \texttt{ACL} $\to$ \texttt{Restaurant} $\to$ \texttt{AI} $\to$ \texttt{Phone}  & 76.49 & 80.84 & {\color{black}75.80}& {\color{black}80.51}& \textbf{76.86} & \textbf{81.09} \\ 
\texttt{Camera} $\to$ \texttt{ACL} $\to$ \texttt{Phone} $\to$ \texttt{Restaurant} $\to$ \texttt{PubMed} $\to$ \texttt{AI} & 76.28 & 80.83 & {\color{black}74.67} & {\color{black}79.91} &
\textbf{77.52} & \textbf{81.65} \\ 
\texttt{AI} $\to$ \texttt{PubMed} $\to$ \texttt{Camera} $\to$ \texttt{Phone} $\to$ \texttt{ACL} $\to$ \texttt{Restaurant} & 76.17 & 80.68 & {\color{black}75.32} & {\color{black}80.45}& \textbf{78.18} & \textbf{82.10} \\
\specialrule{.1em}{.05em}{.05em}
\specialrule{.1em}{.05em}{.05em}
\end{tabular}}
\caption{DAS performance averaged over all domains after the final DAP-trained (averaged over 5 random seeds).} 
\label{tab:order}
\end{table*}

\section{Standard Deviations}
\label{ap:std}
Table~\ref{tab:dapt_std} reports the standard deviations of the corresponding results in Table~\ref{tab:dapt_result} (in the main paper) of DAS and the considered baselines over 5 runs with random seeds. We can see the results of DAS are stable. Some baselines (e.g., RoBERTa in AI, DAP-RoBERTa in Camera) can have quite large standard deviations. 

Table~\ref{tab:ablation_std} reports the standard deviations of the corresponding results in Table~\ref{tab:ablation} (in the main paper) of DAS and the considered baselines over 5 runs with random seeds. We can see the results of DAS and its variants are stable. 

\begin{table*}[]
\centering
\resizebox{\textwidth}{!}{
\begin{tabular}{cc|ccccccccccccc}
\specialrule{.2em}{.1em}{.1em}
\multirow{2}{*}{Category} & Domain & \multicolumn{2}{c}{Restaurant} & \multicolumn{2}{c}{ACL} & \multicolumn{2}{c}{AI} & \multicolumn{2}{c}{Phone} & PubMed & \multicolumn{2}{c}{Camera} & \multicolumn{2}{c}{Average} \\
 & Model & MF1 & Acc & MF1 & Acc & MF1 & Acc & MF1 & Acc & MF1 & MF1 & Acc & MF1 & Acc \\
\specialrule{.1em}{.05em}{.05em}
\multirow{4}{*}{Non-CL} &
{\color{black}Pool}	& {\color{black}$\pm${0.0070}} & {\color{black}$\pm${0.0032}} & {\color{black}$\pm${0.0177 }} & {\color{black}$\pm${	0.0103 }} & {\color{black}$\pm${0.0137 }} & {\color{black}$\pm${0.0087 }} & {\color{black}$\pm${0.0190 }} & {\color{black}$\pm${0.0142 }} & {\color{black}$\pm${0.0088 }} & {\color{black}$\pm${0.0345 }} & {\color{black}$\pm${0.0209 }} & {\color{black}$\pm${0.0127 }} & {\color{black}$\pm${0.0085 }} \\
& RoBERTa & $\pm${0.0117} & $\pm${0.0049} & $\pm${0.0192} & $\pm${0.0096} & $\pm${0.0646} & $\pm${0.0347} & $\pm${0.0210} & $\pm${0.0154} & $\pm${0.0071} & $\pm${0.0403} & $\pm${0.0179} & $\pm${0.0119} & $\pm${0.0070} \\
 & DAP-RoBERTa & $\pm${0.0096} & $\pm${0.0056} & $\pm${0.0218} & $\pm${0.0118} & $\pm${0.0117} & $\pm${0.0086} & $\pm${0.0165} & $\pm${0.0103} & $\pm${0.0035} & $\pm${0.0479} & $\pm${0.0298} & $\pm${0.0118} & $\pm${0.0075}  \\
 & DAP-Adapter &  $\pm${0.0102} & $\pm${0.0068} & $\pm${0.0142} & $\pm${0.0099} & $\pm${0.0551} & $\pm${0.0288} & $\pm${0.0265} & $\pm${0.0181} & $\pm${0.0055} & $\pm${0.0165} & $\pm${0.0110} & $\pm${0.0132} & $\pm${0.0087} \\
 & DAP-Prompt & $\pm${0.0060} & $\pm${0.0035} & $\pm${0.0068} & $\pm${0.0108} & $\pm${0.0301} & $\pm${0.0124} & $\pm${0.0126} & $\pm${0.0087} & $\pm${0.0028} & $\pm${0.0243} & $\pm${0.0138} & $\pm${0.0049} & $\pm${0.0019}\\
\specialrule{.1em}{.05em}{.05em}
\multirow{11}{*}{{\multirow{2}{*}{\begin{tabular}[c]{@{}c@{}}CL\\ DAP-train\end{tabular}}}} & NCL & $\pm${0.0064} & $\pm${0.0035} & $\pm${0.0168} & $\pm${0.0084} & $\pm${0.0164} & $\pm${0.0099} & $\pm${0.0126} & $\pm${0.0104} & $\pm${0.0073} & $\pm${0.0449} & $\pm${0.0247} & $\pm${0.0116} & $\pm${0.0073}\\
 & NCL-Adapter & $\pm${0.0090} & $\pm${0.0060} & $\pm${0.0063} & $\pm${0.0065} & $\pm${0.0835} & $\pm${0.0405} & $\pm${0.0196} & $\pm${0.0124} & $\pm${0.0086} & $\pm${0.0312} & $\pm${0.0152} & $\pm${0.0117} & $\pm${0.0058}  \\
  & DEMIX & $\pm${0.0065} & $\pm${0.0029} & $\pm${0.0118} & $\pm${0.0094} & $\pm${0.0376} & $\pm${0.0218} & $\pm${0.0731} & $\pm${0.0428} & $\pm${0.0069} & $\pm${0.0099} & $\pm${0.0071} & $\pm${0.0121} & $\pm${0.0064}  \\
   & BCL &$\pm${0.0106} & $\pm${0.0059} & $\pm${0.0050} & $\pm${0.0054} & $\pm${0.0433} & $\pm${0.0229} & $\pm${0.0191} & $\pm${0.0130} & $\pm${0.0069} & $\pm${0.0290} & $\pm${0.0164} & $\pm${0.0097} & $\pm${0.0055}  \\
& CLAASSIC & $\pm${0.0071} & $\pm${0.0039} & $\pm${0.0337} & $\pm${0.0171} & $\pm${0.0227} & $\pm${0.0084} & $\pm${0.0187} & $\pm${0.0124} & $\pm${0.0085} & $\pm${0.0140} & $\pm${0.0094} & $\pm${0.0114} & $\pm${0.0065}\\
 & KD & $\pm${0.0352} & $\pm${0.0197} & $\pm${0.0096} & $\pm${0.0107} & $\pm${0.0164} & $\pm${0.0088} & $\pm${0.0149} & $\pm${0.0115} & $\pm${0.0075} & $\pm${0.0277} & $\pm${0.0128} & $\pm${0.0072} & $\pm${0.0042} \\
 & EWC & $\pm${0.0161} & $\pm${0.0085} & $\pm${0.0136} & $\pm${0.0076} & $\pm${0.0178} & $\pm${0.0089} & $\pm${0.0205} & $\pm${0.0140} & $\pm${0.0069} & $\pm${0.0725} & $\pm${0.0424} & $\pm${0.0172} & $\pm${0.0098} \\
& DER++ & $\pm${0.0081} & $\pm${0.0042} & $\pm${0.0156} & $\pm${0.0089} & $\pm${0.0355} & $\pm${0.0160} & $\pm${0.0402} & $\pm${0.0272} & $\pm${0.0090} & $\pm${0.0367} & $\pm${0.0215} & $\pm${0.0158} & $\pm${0.0088} \\
& HAT & $\pm${0.0182} & $\pm${0.0091} & $\pm${0.0271} & $\pm${0.0206} & $\pm${0.0369} & $\pm${0.0126} & $\pm${0.0834} & $\pm${0.0474} & $\pm${0.0038} & $\pm${0.1082} & $\pm${0.0408} & $\pm${0.0323} & $\pm${0.0155} \\
& HAT-All & $\pm${0.0257} & $\pm${0.0140} & $\pm${0.0643} & $\pm${0.0273} & $\pm${0.1355} & $\pm${0.0991} & $\pm${0.0428} & $\pm${0.0217} & $\pm${0.0125} & $\pm${0.0526} & $\pm${0.0163} & $\pm${0.0175} & $\pm${0.0145} \\
 & HAT-Adapter &$\pm${0.0093} & $\pm${0.0061} & $\pm${0.0048} & $\pm${0.0053} & $\pm${0.0289} & $\pm${0.0168} & $\pm${0.0277} & $\pm${0.0195} & $\pm${0.0037} & $\pm${0.0760} & $\pm${0.0370} & $\pm${0.0129} & $\pm${0.0074}  \\
 & DAS &  $\pm${0.0090} & $\pm${0.0063} & $\pm${0.0186} & $\pm${0.0103} & $\pm${0.0142} & $\pm${0.0086} & $\pm${0.0160} & $\pm${0.0135} & $\pm${0.0067} & $\pm${0.0289} & $\pm${0.0154} & $\pm${0.0099} & $\pm${0.0060} \\
\specialrule{.1em}{.05em}{.05em}
\specialrule{.1em}{.05em}{.05em}
\end{tabular}}
\caption{Standard deviations of the corresponding metrics of the proposed
DAS model and the baselines 
} 
\label{tab:dapt_std}
\end{table*}

\begin{table*}[]
\centering
\resizebox{\textwidth}{!}{
\begin{tabular}{cc|ccccccccccccc}
\specialrule{.2em}{.1em}{.1em}
\multirow{2}{*}{Category} & Domain & \multicolumn{2}{c}{Restaurant} & \multicolumn{2}{c}{ACL} & \multicolumn{2}{c}{AI} & \multicolumn{2}{c}{Phone} & PubMed & \multicolumn{2}{c}{Camera} & \multicolumn{2}{c}{Average} \\
 & Model & MF1 & Acc & MF1 & Acc & MF1 & Acc & MF1 & Acc & MF1 & MF1 & Acc & MF1 & Acc \\
\specialrule{.1em}{.05em}{.05em}
\multirow{4}{*}{Non-CL} & RoBERTa & $\pm${0.0117} & $\pm${0.0049} & $\pm${0.0192} & $\pm${0.0096} & $\pm${0.0646} & $\pm${0.0347} & $\pm${0.0210} & $\pm${0.0154} & $\pm${0.0071} & $\pm${0.0403} & $\pm${0.0179} & $\pm${0.0119} & $\pm${0.0070} \\
 & DAP-RoBERTa & $\pm${0.0096} & $\pm${0.0056} & $\pm${0.0218} & $\pm${0.0118} & $\pm${0.0117} & $\pm${0.0086} & $\pm${0.0165} & $\pm${0.0103} & $\pm${0.0035} & $\pm${0.0479} & $\pm${0.0298} & $\pm${0.0118} & $\pm${0.0075}  \\
 & DAP-Adapter &  $\pm${0.0102} & $\pm${0.0068} & $\pm${0.0142} & $\pm${0.0099} & $\pm${0.0551} & $\pm${0.0288} & $\pm${0.0265} & $\pm${0.0181} & $\pm${0.0055} & $\pm${0.0165} & $\pm${0.0110} & $\pm${0.0132} & $\pm${0.0087} \\
 & DAP-Prompt & $\pm${0.0060} & $\pm${0.0035} & $\pm${0.0068} & $\pm${0.0108} & $\pm${0.0301} & $\pm${0.0124} & $\pm${0.0126} & $\pm${0.0087} & $\pm${0.0028} & $\pm${0.0243} & $\pm${0.0138} & $\pm${0.0049} & $\pm${0.0019}\\
\specialrule{.1em}{.05em}{.05em}
\multirow{3}{*}{{\multirow{2}{*}{\begin{tabular}[c]{@{}c@{}}CL\\ DAP-train\end{tabular}}}} & 
{\color{black}DAS (random)}	&
{\color{black}$\pm${0.0074}} & {\color{black}$\pm${0.0055}} & {\color{black}$\pm${0.0110}} & {\color{black}$\pm${0.0102}} & {\color{black}$\pm${0.0201}} & {\color{black}$\pm${0.0112}} & {\color{black}$\pm${0.0184}} & {\color{black}$\pm${0.0128}} & {\color{black}$\pm${0.0042}} & {\color{black}$\pm${0.0483}} & {\color{black}$\pm${0.0247}} & {\color{black}$\pm${0.0119}} & {\color{black}$\pm${0.0067}} \\
& DAS (w/o contrast) & $\pm${0.0104} & $\pm${0.0055} & $\pm${0.0090} & $\pm${0.0063} & $\pm${0.0205} & $\pm${0.0124} & $\pm${0.0321} & $\pm${0.0216} & $\pm${0.0037} & $\pm${0.0527} & $\pm${0.0286} & $\pm${0.0119} & $\pm${0.0073}\\
& DAS (w/o softmask) & $\pm${0.0064} & $\pm${0.0046} & $\pm${0.0121} & $\pm${0.0088} & $\pm${0.0193} & $\pm${0.0113} & $\pm${0.0245} & $\pm${0.0175} & $\pm${0.0096} & $\pm${0.0322} & $\pm${0.0183} & $\pm${0.0104} & $\pm${0.0059} \\
& DAS (w/o initialization) & $\pm${0.0124} & $\pm${0.0075} & $\pm${0.0054} & $\pm${0.0048} & $\pm${0.0134} & $\pm${0.0078} & $\pm${0.0135} & $\pm${0.0104} & $\pm${0.0118} & $\pm${0.0460} & $\pm${0.0261} & $\pm${0.0093} & $\pm${0.0058} \\
& DAS (domain-specific) & $\pm${0.0067} & $\pm${0.0045} & $\pm${0.0151} & $\pm${0.0127} & $\pm${0.0192} & $\pm${0.0129} & $\pm${0.0277} & $\pm${0.0182} & $\pm${0.0061} & $\pm${0.0419} & $\pm${0.0226} & $\pm${0.0120} & $\pm${0.0077} \\
 & DAS &  $\pm${0.0090} & $\pm${0.0063} & $\pm${0.0186} & $\pm${0.0103} & $\pm${0.0142} & $\pm${0.0086} & $\pm${0.0160} & $\pm${0.0135} & $\pm${0.0067} & $\pm${0.0289} & $\pm${0.0154} & $\pm${0.0099} & $\pm${0.0060} \\
\specialrule{.1em}{.05em}{.05em}
\specialrule{.1em}{.05em}{.05em}
\end{tabular}}
\caption{Standard deviations of the corresponding metrics of the proposed
DAS model and the ablation 
} 
\label{tab:ablation_std}
\end{table*}

\end{document}